\documentclass[10pt,journal,compsoc]{IEEEtran}

\ifCLASSOPTIONcompsoc
  \usepackage[nocompress]{cite}
\else
  \usepackage{cite}
\fi
\usepackage{hyperref}
\ifCLASSINFOpdf
  \usepackage[pdftex]{graphicx}
  \usepackage{amsmath}
  \usepackage{subfigure}
\else

\fi

\usepackage{amsmath}

\usepackage[ruled]{algorithm2e}

\usepackage{multirow}
\usepackage{multicol}
\usepackage{threeparttable}

\newcommand{\etal}{\textit{et al}. }

\begin{document}
\title{Correlation Filter Selection for Visual Tracking Using Reinforcement Learning}

\author{Yanchun Xie, Jimin~Xiao, ~\IEEEmembership{Member,~IEEE}, Kaizhu~Huang,~\IEEEmembership{Member,~IEEE},
Jeyarajan~Thiyagalingam, ~\IEEEmembership{Member,~IEEE},  Yao~Zhao,~\IEEEmembership{Senior Member,~IEEE}

\thanks{This work was supported by the National Natural Science Foundation of China under Grant 61501379 and Grant 61210006 and by
the Jiangsu Science and Technology Programme (BK20150375). (Corresponding author: Jimin Xiao.)}
\thanks{Y.~Xie, J.~XIAO, K.~Huang  are with the Department of
Electrical and Electronic Engineering, Xi'an Jiaotong-Liverpool University,
111 Ren Ai Road, Suzhou, P.R. China (e-mail: yanchun.xie, jimin.xiao, kaizhu.huang@xjtlu.edu.cn).}
\thanks{J.~Thiyagalingam is with Science and Technologies Facilities Council, Rutherford Appleton Laboratory, Oxon, UK (e-mail: t.jeyan@stfc.ac.uk).}
\thanks{Y.~ZHAO is with Institute of Information Science, Beijing Jiaotong University,
Beijing, China (e-mail: yzhao@bjtu.edu.cn).}
}

\IEEEtitleabstractindextext{%
\begin{abstract}
Correlation filter has been proven to be an effective tool for a number of approaches in visual tracking, particularly for seeking a good balance between tracking accuracy and speed. However, correlation filter based models are susceptible to wrong updates stemming from inaccurate tracking results. To date, little effort has been devoted towards handling the correlation filter update problem. In this paper, we propose a novel approach to address the correlation filter update problem. In our approach, we update and maintain multiple correlation filter models in parallel, and we use deep reinforcement learning for the selection of an optimal correlation filter model among them. To facilitate the decision process in an efficient manner, we propose a decision-net to deal target appearance modeling, which is trained through hundreds of challenging videos using proximal policy optimization and a lightweight learning network. An exhaustive evaluation of the proposed approach on the OTB100 and OTB2013 benchmarks show that the approach is effective enough to achieve the average success rate of  62.3\% and the average precision score of  81.2\%, both exceeding the performance of traditional correlation filter based trackers.

\end{abstract}

}

\maketitle

\pagestyle{empty}  % no page number for the second and the later pages
\thispagestyle{empty} % no page number for the first page

\IEEEdisplaynontitleabstractindextext

\IEEEpeerreviewmaketitle

\IEEEraisesectionheading{\section{Introduction}\label{sec:introduction}}

Visual object tracking is a process of locating objects of interest precisely over a sequence of image frames, given a bounding box in the initial frame. Instance-level discrimination plays a vital role in visual tracking. Other than object recognition tasks, an accurate object tracker should be able to distinguish not only generic objects from the background but also recognize and differentiate them from similar objects. To this end, handling objects that are of similar in color or geometry to the objects of interest --- distractor objects, is a key challenge during the feature extraction stage of the visual tracking.

Discriminative correlation filter (CF)-based trackers~\cite{bolme2010visual}\cite{danelljan2014accurate}\cite{henriques2012exploiting}\cite{li2014scale}\cite{henriques2015high} achieve a good trade-off between accuracy and speed by efficiently solving a ridge regression problem in Fourier frequency domain. Regularized correlation filters~\cite{danelljan2015learning}\cite{danelljan2016adaptive} are proposed to further enhance the tracking accuracy. Gladh~\etal introduces motion information along with hand-crafted features for CF tracking~\cite{gladh2016deep}. Mueller~\etal propose a context-aware CF tracking~\cite{mueller2017context}. Sophisticated learning schemes are proposed to achieve powerful feature representation~\cite{danelljan2017eco}\cite{song2017crest}.

Most discriminative model-based trackers exploit the target from a given bounding box directly, which is used to build the appearance model of the objects at latter stages. During the tracking process, new image patches generated from new frames are supplemented to further update the CF model. Generally, a small update-rate is usually preferred for CF trackers in order to maintain model stability. These trackers may easily suffer from a drift problem, especially in challenging environments such as partial occlusions, background clutter, and low resolution.

\begin{figure}[h]
\centering
\includegraphics[width=1\linewidth]{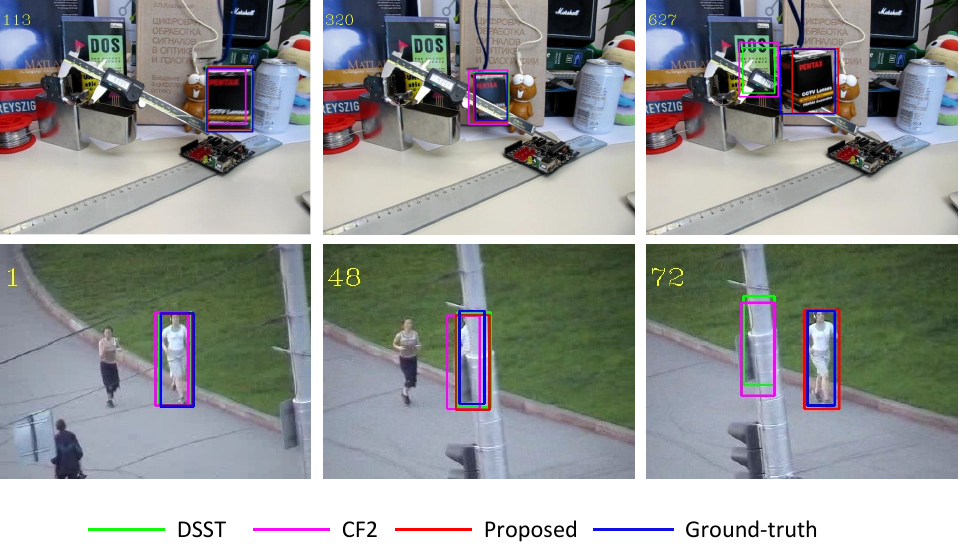}\\%{structure.jpg}
\caption{Visualization of 3 tracking results. Green, purple, red box denote tracking results of DSST, CF2, and the proposed tracker,
respectively;  blue box denotes the ground-truth box of tracking sequences. During the tracking process, targets suffer from partial occlusion, while other trackers do not realize this and result in model drift. The proposed tracker with decision unit updates the appearance model guided by the response map and skips updating if not necessary. }

\label{fig1}
\end{figure}

An example of this is illustrated in Fig.~\ref{fig1}, the tracking model is initialized with a target box in the first frame, which is also used as the ground truth for subsequent analyses. A discriminative CF can easily be obtained using a two-dimensional Gaussian label whose center is the same as that of the target box. However, during the tracking process, we notice the target becomes partially occluded by foreground objects in some cases, as shown in the middle column of Fig.~\ref{fig1}. However, the CF model is oblivious to this occlusion issue and kept updated without even evaluating the reliability of new image patches. Tracking results that are generated with such poorly updated CF models influence the subsequent updates of the CF model. A number of such updates accumulate the errors and results in irrecoverable model drift.

To mitigate such model drifts, Gao~\etal propose a deep network to learn a relative model to deal with target appearance changes~\cite{gao2017deep}. Yao~\etal propose a semantics-aware method~\cite{yao2018semantics} to enhance appearance model in visual object tracking. However, it is not flexible to transfer a relative model or add semantics information into a CF-based tracker. Furthermore, such a transfer process entails substantial investment in time towards re-modeling the relative model. Different from ensembling siamese network~\cite{jiang2018siamese}, a decision-making network is proposed using a Siamese tracking framework~\cite{choi2017visual}, which also aims to solve the model drifting problem. Also in other tasks like persion search~\cite{xiao2018ian} which aims to find a more discriminative feature to handle huge variance of visual appearance. Base on that, we naturally consider that a selection for CF models will contribute to building a better discriminative appearance model for visual tracking.

In recent years, progress in deep learning has been influential in the domain of visual tracking. Convolutional features has been considered in several studies~\cite{ma2015hierarchical,danelljan2015convolutional, danelljan2016beyond, song2017crest, wang2017dcfnet, chen2017once}. These studies show that deep convolutional networks (DCNs) that are pre-trained 
with certain large-scale data and adaptive correlation filter are complementary. The CF-embedded DCNs are shown to be able to achieve state-of-the-art performance on many object tracking benchmarks~\cite{wu2013online}.

An approach of CF-based tracking reformulating the CF into a convolutional layer can offer end-to-end learning.
For example, in~\cite{song2017crest}, instead of solving the CF with a closed-form solution, it is learned as kernels of a convolutional layer, which can benefit from end-to-end training. In this framework, the CF is updated by back-propagation. However, despite using residual learning to enhance the feature representation, noisy updates are still a problem. Meanwhile,  the application of Siamese frameworks has also been explored in visual tracking, including SiameseFC~\cite{bertinetto2016fully}, DSaim~\cite{guo2017learning}, SINT~\cite{tao2016siamese} and CFNet~\cite{valmadre2017end}. They all employ powerful convolutional network to address the similarity learning problem for visual tracking.

Although the utilization of both convolutional neural network (CNN) and CF have been instrumental in addressing a number of problems and in achieving rather remarkable outcomes, there are still a number of problems still remain to be addressed.

First,  when obtaining discriminative features for tracking, owing to the underlying complexity of parameter models,  significant amount of computational resources are needed. In addition to this,  large models tend to introduce severe over-fitting problems. Models like VGG-19 tend to be an inferior option for CF-based trackers. Other than one forward pass in the convolutional network for feature extraction, CF trackers need additional time to compute the correction filter in the Fourier frequency domain which can hardly benefit from GPUs. Nevertheless, operating in the Fourier frequency domain speeds up CF.

Second, most existing trackers update tracking models at each frame. Especially for CF trackers, a simple moving average scheme is exploited in essence.
For example, the state-of-the-art tracker ECO~\cite{danelljan2017eco} takes the sparser update to refine their model. This may, however, cause deterministic failures once the target is inaccurately detected, severely occluded or totally missing in the current frame. Meanwhile, it is hard to judge whether an update for the CF is reliable or not. Therefore, a more sophisticated model update strategy is necessary to handle this issue.

Motivated by the fact that the CF model might be updated with inaccurately tracking results, some temporally old CF models
might be able to generate better tracking results than the latest one. In this paper, we propose to maintain more than one CF model.
Instead of always using the latest CF model, the most suitable CF model will be selected and used to generate tracking results.
To select the most suitable one among multiple models, reinforcement learning is deployed.

Convolutional features contribute to robust feature representation. Therefore, in our proposed method,
we engage a light-weighted convolutional network as feature extractor.
Meanwhile, the performance of CF-based trackers, in comparison to other trackers,  is a great advantage. While a standard CF solver is exploited for tracking, the net structure in~\cite{wang2017dcfnet} satisfies the need for fast convolutional feature extraction.
Based on this work, we investigate the model update problem by formulating CF model updating as a Markov decision process.

Reinforcement learning has been studied for visual tracking recently~\cite{huang2017learning}, ~\cite{yun2017action}.
Huang~\etal~\cite{huang2017learning} succeeds in utilizing Q-learning~\cite{mnih2015human} for shallow-level or high-level feature selection.
ADnet~\cite{yun2017action} uses policy gradient learning and trains action dynamics for tracking with annotated visual tracking sequences. Recently, Dong ~\etal ~\cite{dong2018hyperparameter} propose to use continuous deep Q-Learning for hyperparameter selection in tracking.
Our work is significantly different from these existing works, in that we are studying the model update issue with reinforcement learning.

The main contributions of this paper are as follows:

\begin{enumerate}

 \item  We propose a novel approach for selecting an optimal model among multiple CF models which are updated and maintained in parallel. This approach addresses a number of concerns that arise from a single CF model, such as drift;

\item We propose a reinforcement learning-based approach for optimal model selection. To the best of our knowledge, this is the first time that reinforcement learning is utilized for model selection among multiple CF models;

\item We utilize a light-weight feature extractor and proposed a small decision network so that the proposed approach can be deployed  in real-time applications, where the frame rates are high;

\item We exhaustively evaluate the proposed approach on OTB100 and  OTB2013 benchmarks. Our results show an average success rate of  $62.3\%$ and average precision $81.2\%$. These results are better than the approaches that adopt traditional CF trackers without multiple model selection.

\end{enumerate}

The rest of this paper is organized as follows. In Section~\ref{sec2}, we present a detailed literature survey. The proposed tracking method with implementation details is described in Section~\ref{sec3}. We present and discuss the experimental results in Section~\ref{sec4}. Finally, conclusions are set out in Section~\ref{sec5}.

\section{Related Work}
\label{sec2}
%history of cf
\subsection{Correction Filter Based Tracker}

CF-based trackers achieve a good trade-off between accuracy and speed by solving a ridge regression problem efficiently in the Fourier frequency domain.
After Bolme~\etal introduced the CF for fast visual tracking, several bodies of work have been proposed to improve the tracking performance of CF-based approaches. Henriques~\etal  propose a circulant structure kernel tracker (CSK)~\cite{henriques2012exploiting}. A high-speed tracker with kernelized correlation filters (KCF) is proposed in~\cite{henriques2015high}. In KCF~\cite{henriques2015high},
a multi-channel Histogram Of Gradient (HOG) feature is introduced to calculate the CF. Danelljan~\etal introduce a scale pyramid representation~\cite{danelljan2014accurate} to handle the scaling issue and proposed the 3-dimensional CF. In~\cite{danelljan2017discriminative}, separate discriminative correlation filters were
learned for translation and scale estimation, respectively.
To mitigate unwanted boundary effects, Danelljan~\etal introduced a spatially regularized term~\cite{danelljan2015learning} to penalize CF coefficients based on their spatial locations.
Unfortunately, the improvement in accuracy goes along with significant reductions in tracking speed.

Some other CF methods focus on improving the feature representation by directly taking several layers of a pre-trained deep network
like VGG~\cite{ma2015hierarchical,danelljan2015convolutional}.
On top of pre-trained convolutional layers, convolution operator tracker (COT)~\cite{danelljan2016beyond} was proposed to integrate multi-resolution convolutional features in different layers. The CREST~\cite{song2017crest} framework reformulated the CF into a convolutional layer. In addition,  Qiang ~\etal present an end-to-end light-weight network architecture~\cite{wang2017dcfnet} to learn better features that fit the CF model using off-line training. In their work, a CF is treated as a special layer added to a Siamese network. Feature extractor consisting of two convolutional layers is trained for the online tracking task. We exploit the feature extractor from~\cite{wang2017dcfnet} and further investigate the CF model update problem using the latest reinforcement learning algorithms~\cite{schulman2015trust},~\cite{schulman2017proximal}.

\subsection{Deep Reinforcement Learning}

Deep reinforcement learning (RL) algorithms have already been applied to various problems arising from different domains. Control policies for robots can be learned by RL directly from real camera outputs~\cite{levine2016end,levine2016learning}. Deep learning enables RL to scale to decision-making problems. The standout success of AlphaGo, which defeated a human world champion in Go, has shown deep RL can handle complex states and action spaces very well. Also deep RL is applied for many computer vision tasks like objection localization~\cite{caicedo2015active},~\cite{jie2016tree}, object detection \cite{mathe2016reinforcement}, action recognition~\cite{tang2018deep} and person re-identification~\cite{zhang2017multi}.

High variance in gradients makes it difficult to train a deep RL network.
Actor-critic methods~\cite{silver2014deterministic,mnih2016asynchronous} utilize learned value function as feedback term to guide the training.
Trust region policy optimization (TRPO) \cite{schulman2015trust} has been shown to be relatively robust and applicable to domains with high-dimensional inputs.
To achieve this, TRPO optimizes a surrogate objective function, specifically, it optimizes an (importance sampled) advantage estimate, constrained with a quadratic approximation of KL divergence.
The latest proximal policy optimization (PPO) \cite{schulman2017proximal} algorithm performs unconstrained optimization, requiring only first-order gradient information.
Due to its good performance, PPO is gaining popularity for a range of deep RL tasks.
Our work also uses the PPO algorithm to learn a policy for selecting an appropriate CF model for visual tracking.

Employing the deep RL algorithms into computer vision problems could benefit from the experience.
In fact, RL has been studied for visual tracking in several recent works~\cite{huang2017learning,yun2017action},~\cite{zhang2017deep},~\cite{supancic2017tracking}.
Huang~\etal succeed in utilizing Q-learning \cite{mnih2015human} for shallow-level or high-level feature selection~\cite{huang2017learning}.
Yun~\etal propose an action-decision network\cite{yun2017action} used policy gradient learning and trained action dynamics for tracking with annotated visual tracking sequences. Luo~\etal propose an active tracking scheme trained in simulators by reinforcement learning~\cite{luo2018end}.
Our work distinct from these existing works, in that we are studying the model update issue with reinforcement learning.

\begin{figure}[!h]
\centering
\includegraphics[width=0.8\linewidth]{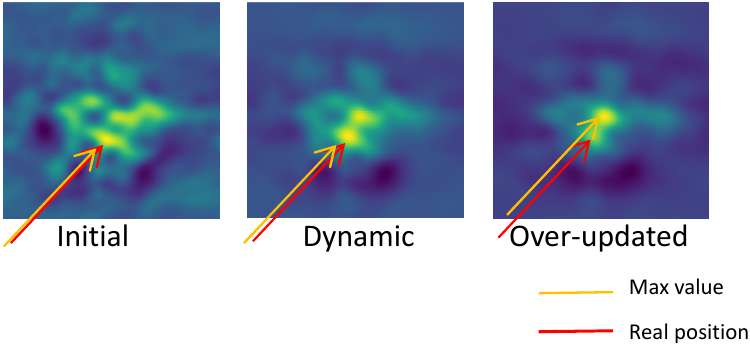}%{structure.jpg}
\caption{A visualization of 3 response maps from CF models of different stages. Bright yellow color denotes regions where high probability the target will be, while the dark blue color represents a relatively low probability. After a period of update, CF model drifts and the over-updated model produces a good-looking response map while failing to track the true target. (Better viewed in color)}

\label{fig2}
\end{figure}

\begin{figure*}[ht]
\centering
\includegraphics[width=0.85 \linewidth]{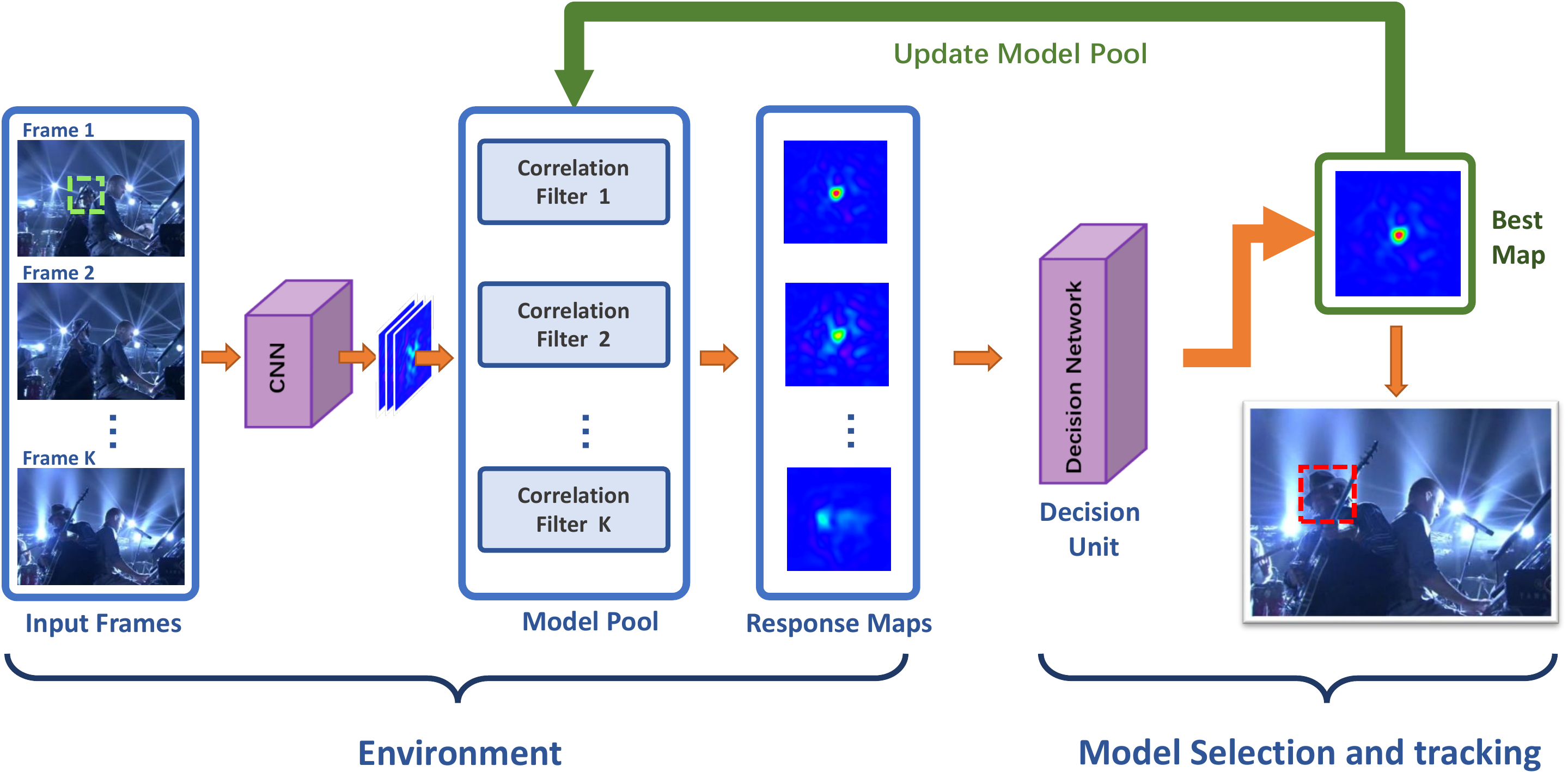}%{structure.jpg}

\caption{Given a sequence containing $L$ frames, we take the target location in the first frame and use its feature to initialize the CF model. Different states at each time-step would produce CF models sharing different memories of target appearance.  From each CF model, we obtain one corresponding response map. The trained decision network will select the response map based on learned experience and finally point out the target locations.}

\label{fig3}
\end{figure*}
\section{Our Approach}
\label{sec3}
%\IEEEpubidadjcol

In visual tracking, the traditional CF model might be updated with inaccurately tracking results, and suffers from drift problem, as shown in Fig.~\ref{fig1}.
To mitigate the issue of possible inaccurate model update
during the tracking process, we propose to maintain more
than one CF model for visual tracking. Instead
of always using the latest CF model, the most suitable CF model will be selected using reinforcement learning.
More specifically, the current search frame is input into the convolutional feature extractor,
and several response maps are generated utilizing all the maintained CF models.
Each response map corresponds to one CF model.
Different response maps at one-time step are visualized in Fig.~\ref{fig2}.
The RL algorithm PPO~\cite{schulman2017proximal} is utilized to
select the most suitable CF model based on the convolutional feature of the corresponding response map.
Then, the tracking bounding box is generated using the corresponding CF response map.
Finally, the CF models are updated with the new tracking bounding box, which will be used for the next frame.
The overall proposed framework is presented in Fig.~\ref{fig3}.
The pseudo-code of the algorithm is described in Algorithm~\ref{algorithm1}.

\begin{algorithm}
\caption{Visual tracking with multiple CF models and reinforcement learning.}
\KwIn{\\Tracking sequence of length $L$\\
       Initial object location $x_0$}
\KwOut{Target object location in frame t $x_t$}

\vspace{2ex}

Initialize CF model $M_0$ with the ground-truth \\
Set history CF model $M_1$, .. , $M_{K-1}$ = $M_0$ \\
\For{$t=1$ to $L$}
{
\For{$i=1$ to K}
{
Calculate response maps $P_i$ with each $M_i$\\
Produce confidence score via decision net $\pi(a_t|s_t;\theta)$ for each $P_i$
}
Choose the response map with maximum confidence $P_m$ ;\\
Localize the target according to chosen $P_m$;\\
Update corresponding CF models and save to history;
}
\label{algorithm1}
\end{algorithm}

In this section, we first present the standard problem formulation for CF-based tracking.
Then, we introduce the decision-making process using reinforcement learning.
Finally, we explain how to train the RL model and design the new CF update mechanism.

\subsection{Correlations Filter Model}

%ADD CF algorithms.
The CF-based trackers have demonstrated strong capability on building accurate models with slight online model updating.
Recently, many proposed new tracking algorithms~\cite{song2017crest,galoogahi2017learning}
benefit from the advantage of CF. A standard CF can be solved following the objective
function (\ref{equ_cf}),
\begin{equation}
 \underset{f}{\text{arg min}} =|| \psi(x)*f - g ||^2 + \lambda||f||^2 ,
\label{equ_cf}
\end{equation}
where $f$ is the CF, $*$ is the circular correlation or convolution operation, and  $\psi$ is a feature extractor, $x$ is a
cropped image centered on the target, and $g\in R^{H\times W}$ is the desired Gaussian shaped response map label.
 $f$ can be efficiently solved by transforming (\ref{equ_cf}) into the Fourier domain.
The Fourier domain representation of $f$ can be calculated as (\ref{equ_fourier}).
\begin{equation}
 F =\frac{\bar{G} \odot \bar{X}}{\bar{X} \odot X+\lambda} ,
\label{equ_fourier}
\end{equation}
where $G$ is the Fourier transformation from Gaussian shaped label $g$, $X$ is the Fourier transformation of $x$, and the bar means complex conjugation. Operator $\odot$ is the element-wise product.

New search image $z$ around the target in the next frame is cropped with 2 to 4 times of the target size.
A response map $P$ in the Fourier domain is obtained by (\ref{equ_fourier1}).
\begin{equation}
 P = F \odot \bar{Z} ,
\label{equ_fourier1}
\end{equation}
where $Z$ is the Fourier transformation of $z$.
At a new tracking frame, once the CF $F$ is ready,
the tracking bounding box center locates at the coordinate that has the maximum response value.

Typically, the numerator A and denominator B of the CF in (\ref{equ_fourier}) are updated separately using a moving average mechanism.
\begin{eqnarray}
A_t = (1 - \eta) A_{t-1} + \eta G * \bar{X_t} ,\\
B_t = (1 - \eta) B_{t-1} + \eta X_t * \bar{X_t} + \lambda ,
\end{eqnarray}

Traditional CF trackers update tracking models frame by frame without considering their tracking results.
This may cause an inaccurate model update when occlusion or object missing occurs.
Designing a criterion to produce high-confidence update has been exploded by \cite{wang2017large}. Average peak-to-correlation energy (APCE) is proposed to select high-confidence response maps which effectively prevent CF model from corruption.
In this paper, instead of calculating an APCE score to decide whether to update the model or not,
we introduce a learning algorithm to perform multiple model selection.

%\section{Paper formatting}
%

\begin{figure*}[ht]
\centering
\includegraphics[width=1 \linewidth]{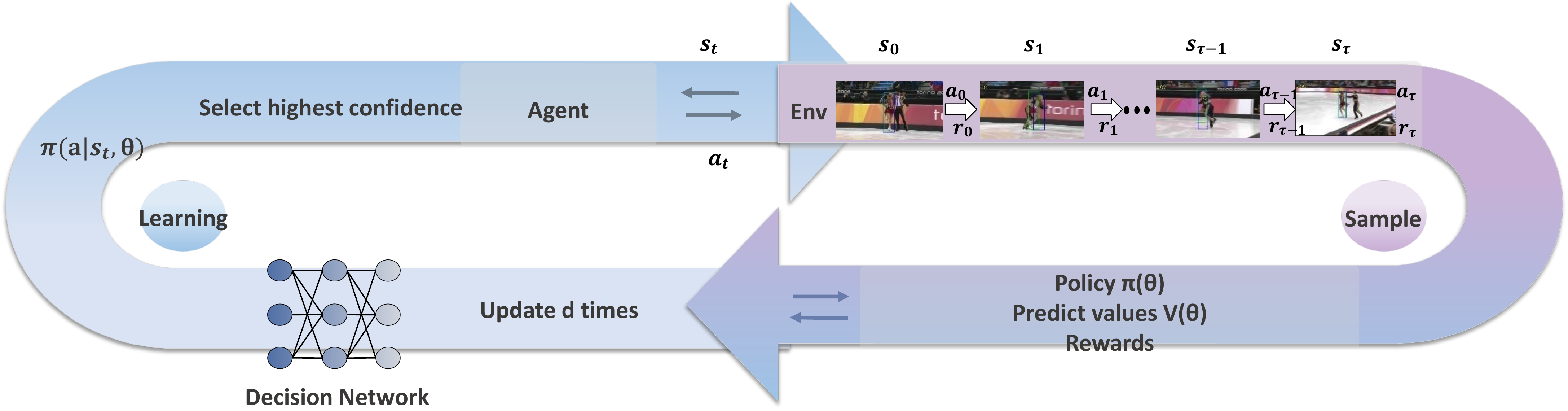}%{structure.jpg}

\caption{Training process of reinforcement learning algorithm for tracking. Decision network which consists of a policy network and a value network, takes the observation from the environment and produce instructions for the agent to act. $a_t$ here is to select appropriate CF models which generate response map to speculate target location. A collection \(\{(s_0, a_0, r_0), \cdots ,(s_t, a_t, r_t), \cdots , (s_T, a_T, r_T)\}\) is sampled after a series of actions, which is used to update the decision network. }
\label{fig_structure}
\end{figure*}

%\subsubsection{Response map}

\subsection{Model Selection Using Reinforcement Learning}

We formulate object tracking as a discrete control problem, which requires the tracker to rapidly respond to object's movement and appearance change based on CF response maps.

In the RL set-up, the agent interacts with the environment by taking an action corresponding to the current state. After the agent receives a state, the agent uses its policy to take an action. Both the environment and the agent will transit to a new state based on the current state and the chosen action. A reward evaluating the made action will be used as feedback and sent to the decision unit to learn and improve the policy.

At frame $t$, we denote the observed state by \(s_t\), which is a set of response maps generated by the CF models.
Denote the action by $\mathcal{A}$ of size $k$, which represents selecting $k$ different CF models.
At each frame, we draw an action $a_t$ $(a_t \in \mathcal{A})$, from a policy distribution.
Then, a reward, $r_t$, according to the tracking results, can be calculated and obtained after the agent's action.
The reward is computed through reward function $r_t = g(s_t, a_t)$, and we will detail the function later.
The old state is updated by the agent and a new state $s_{t+1}$ will be generated which is an unknown state depending on the taken action.
Repeating this process, we can observe a sequence of \{state, action, reward\}, denoted as  $\tau  = \{(s_0, a_0, r_0), \cdots ,(s_t, a_t, r_t), \cdots , (s_T, a_T, r_T) \}$.
Here, at time-step $T$, the tracker reaches the end of the sequence or it fails to locate position inside the
image.
The collected samples are used to update the decision network.
The block diagram of reinforcement learning for visual tracking is illustrated in Fig.~\ref{fig_structure}.

Meanwhile, we can learn policy function $\pi(s_t;\theta)$ and value function $V(s_t;\theta)$ over the trace $\tau$ with stochastic policy gradient and value function regression using PPO \cite{schulman2017proximal}. The loss function $L_t(\theta)$ is defined as follows, which combines the  policy surrogate and value function term.
\begin{equation}
L_t(\theta) = min ( \ Ratio * A_t,\; clip( Ratio ,1 - \epsilon,\; 1 + \epsilon)\; A_t ) ,
\label{equ_loss}
\end{equation}
where
\begin{equation}
Ratio = \frac{\pi(a_t|s_t;\theta)}{\pi(a_t|s_t;\theta_{old})} ,
\end{equation}
Here, $\theta_{old}$ is the vector of policy parameters before the update.
$\theta$ is the new policy parameters. $\pi(a_t|s_t;\theta)$ is the policy function, which defines
the probability to take action $a_t$, under the state $s_t$ and policy parameters $\theta$.
Similarly, $\pi(a_t|s_t;\theta_{old})$ is the the probability to take action $a_t$, under the state $s_t$ and the old policy parameters $\theta_{old}$.

The clipped surrogate objective limits the variation of the surrogate, which adds constraint between the old and new policy before and after the update.
Parameters will be updated based on the collected $\tau$ in time when $T$ time-step is over. Adam optimizer is used for updating the policy and value network. $\epsilon$ is the clipping parameter which is set to $0.2$ in this paper.

$A_t$ is the advantage estimation given state $s_t$, which includes both the current and future rewards.
\begin{equation}
A_t = r_t + \gamma r_{t+1}\ + .. + \gamma^{T-t+1}  r_{T-1}  + \gamma^{T-t}  r_{T}  - V(s_t;\theta) ,
\end{equation}
Here $A_t$ is the difference between the accumulated reward and the estimated state value $V(s_t;\theta)$.
In actor-critic algorithms, the advantage function is the difference between
the accumulated reward and the estimated average reward, defined as value function  $V(s_t;\theta)$.

\vspace{2ex}
\noindent
\textbf{State and Action}
Generally, the state comprises sufficient information from the environment for the agent to take actions. Other than directly taking the input image patches as the state like ADnet \cite{yun2017action}, in our proposed method,
all the response maps produced by corresponding CF models are used as the state.
In the proposed visual tracking framework, an action is defined to select one response map among all candidates by the agent.
Actions are sampled from a policy distribution $\pi$, and the action with the highest score is more likely been chosen by the agent.
The selected response map is used to generate tracking results in the current frame, which will be used to update CF models.
These updated CF models will generate response maps in the following frames,
which serve as the state of the next time slot. This above state transition process will repeat until the last frame.

\noindent
\textbf{Reward}
%########################################################################################## Mod 
The reward function is defined as \(r_t = g(s_t, a_t)\).
A total accumulated reward can be produced until the termination time-step $T$.
At termination time-step $T$, the tracker reaches the end of the sequence or it fails
to locate position inside the image.
%########################################################################################## Eod 
\begin{equation}
g(s_t, a_t) =\left\{
\begin{aligned}
& & IOU + 1 & & IOU>0.7\\
& & -1      & & IOU<0.2\\
& & -0.1    & & otherwise 
\end{aligned}
\right. ,
\end{equation}
where IOU denotes the overlap ratio between tracking result and the ground-truth.

\begin{figure}[!h]
\centering
\includegraphics[width=0.9\linewidth]{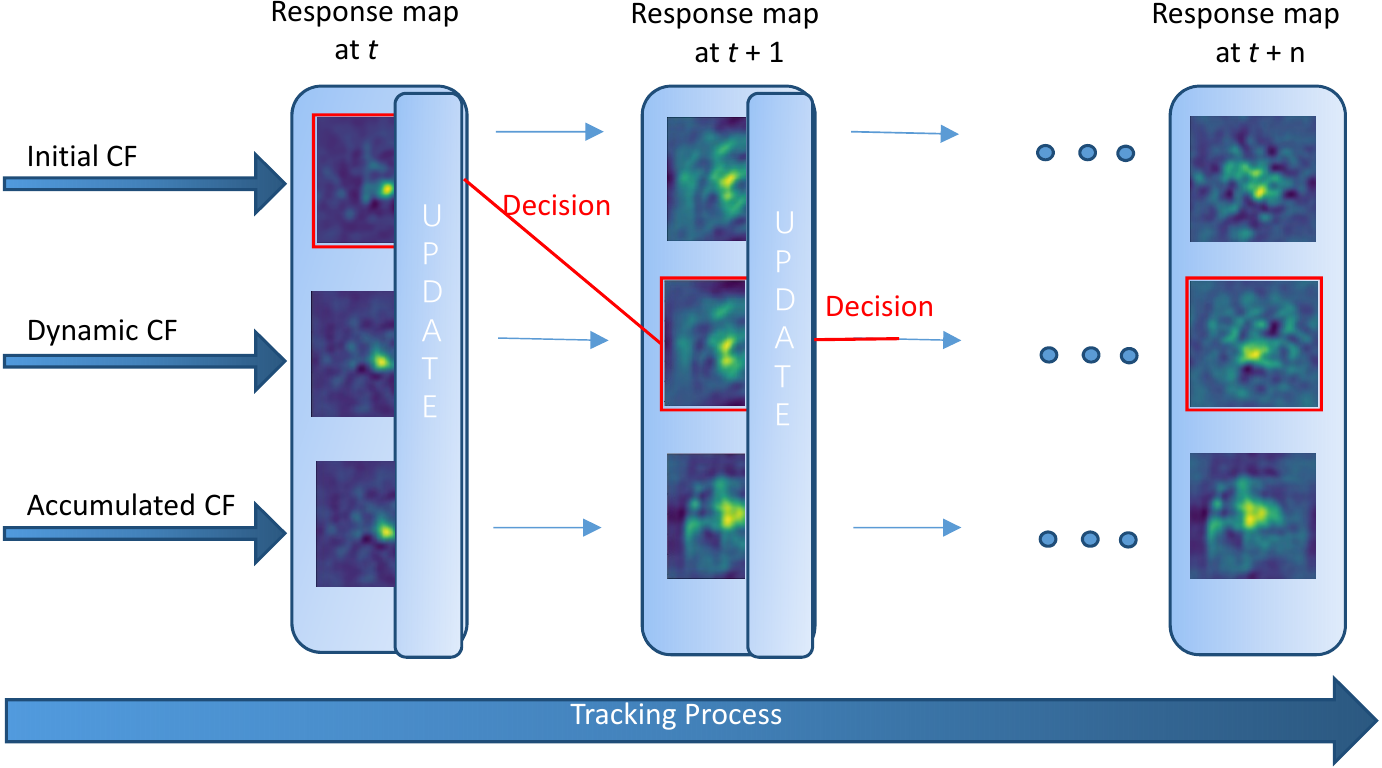}
\caption{Decision making in the tracking process. CF models numbers shown in the example is $k=3$. As shown in this figure, an initial model is chosen at time $t$ which result in the dynamic model unchanged until the time $t+1$ while only the accumulated model is updated. After that, both the dynamic model and the accumulated model are updated at time $t+1$ because of the activation of the dynamic model.  }
\label{fig_dec}
\end{figure}

\noindent
\textbf{Correlation Filter Update}
In order to build better discriminative appearance model, we keep $k$ CF model in our framework, including 1 initial model, 1 accumulated model, and $k-2$ dynamic models.
Siamese trackers only compare the difference between candidates in search image and the ground-truth in the first frame.
So we continue to have the initial CF in our model pool without any updates.
Model drift would easily happen when the tracker lost the memory of the original targets.
Also, we always keep another accumulated CF model in our model pool in order to better adapt to the viewpoint change, deformation and other variations. Between these two typical situations, $k-2$ dynamic CFs play the role of 'peacemaker' and the update for dynamic CFs only activated when chosen by the decision net.
All the $k$ models are initialized by the given tracking target, and the dynamic models are adaptively updated, during the tracking process.
The decision-making process of our approach is shown in Fig.~\ref{fig_dec}.

\subsection{Decision Network}

The response maps are input to the decision network for selection. The network includes $2$ branches, the policy net branch, and the value net branch using the actor-critic framework.
As described in Table~\ref{table_net}, the $2$ branches have separate convolutional layers, one shared fully connected layer and another separate
fully connected layers.

Response maps of a new input frame are resized to $64 \times 64 \times 3$  image and fed to the network as input,
and here we call it the state or observation.
Then the policy net branch will produce a distribution over all actions.
It is worth noticing that action probability distribution is generated through beta distribution~\cite{chou2017improving}.
Finally, an action with the highest probabilities is selected.

Unlike the policy gradient algorithm for online adaptation in ADnet~\cite{yun2017action}, we adopt the actor-critic framework.
An expected accumulated reward is generated by the value function for one specific policy, which guides the "actor" (policy) to learn by taking feedback from the "critic" (value function) and reduces the variance of policy gradient during the training.

\begin{table}[!h]
\begin{center}
\begin{threeparttable}
\caption{The structure of our decision network.
 }
\scriptsize
\begin{tabular}{|c|c|c|c|c|c|}
\hline
Layers  &  \#1 &   \#2   &   \#3  &  \#4  & \#5   \\
%& (VGGNet) & (ResNet-50) & (ResNet-101) & (ResNet-50) & (ResNet-50) & (ResNet-101) \\
\hline
\multirow{2}{*}{Policy net}  & $C5 \times 5$ &  $C3 \times 3$ & $C3 \times 3$  & \multirow{4}{*}{FC512} & \multirow{2}{*}{FC512} \\
                             & $-32S2$       &  $-32S2$       & $-32S2$        &                        &    \\
\cline{1-4}

\multirow{2}{*}{Value net}   & $C5 \times 5$ &  $C3 \times 3$ & $C3 \times 3$  &                        & \multirow{2}{*}{FC512}\\
                             & $-32S2$       &  $-32S2$       & $-32S2$        &                        &    \\
\hline
\label{table_net}
\end{tabular}
\begin{tablenotes}
\item[1]  $C5 \times 5-32S2$ means 32 filters of size $5 \times 5$ and stride 2. FC512 indicates dimension 512.
\end{tablenotes}
\end{threeparttable}
%\caption{where $C5 \times 5-32S2$ means 32 filters of size $5 \times 5$ and stride 2. FC512 indicates dimension 512.}

\end{center}

\end{table}
\subsection{Reinforcement Training with PPO}

\noindent

\textbf{Environment Setup} To avoid over-fitting, we used a large-scale video detection dataset VID~\cite{ILSVRC15} for training our tracker. VID consists of $30$ object categories, which is a subset of $200$ categories in the object detection dataset. We sub-sampled the dataset and choose videos whose target size is less than $60\%$ of their frame size.

To improve the training efficiency, we first test all selected videos via CF tracker.
Based on the tracking accuracy, all the sequences are classified into 3 categories, including easy sequence, extremely hard sequences, and moderate sequences~\cite{shi2016embedding}. We exclude easy and extremely hard sequences from the training set,  since (1) easy sequences will produce $k$ similar good responses maps that vague the decision criterion, and (2) those extremely hard sequences can provide less valid samples and ambiguous labels for RL training.

\begin{figure*}[htb]
\centering
\subfigure[]{
\includegraphics[width=0.24\linewidth]{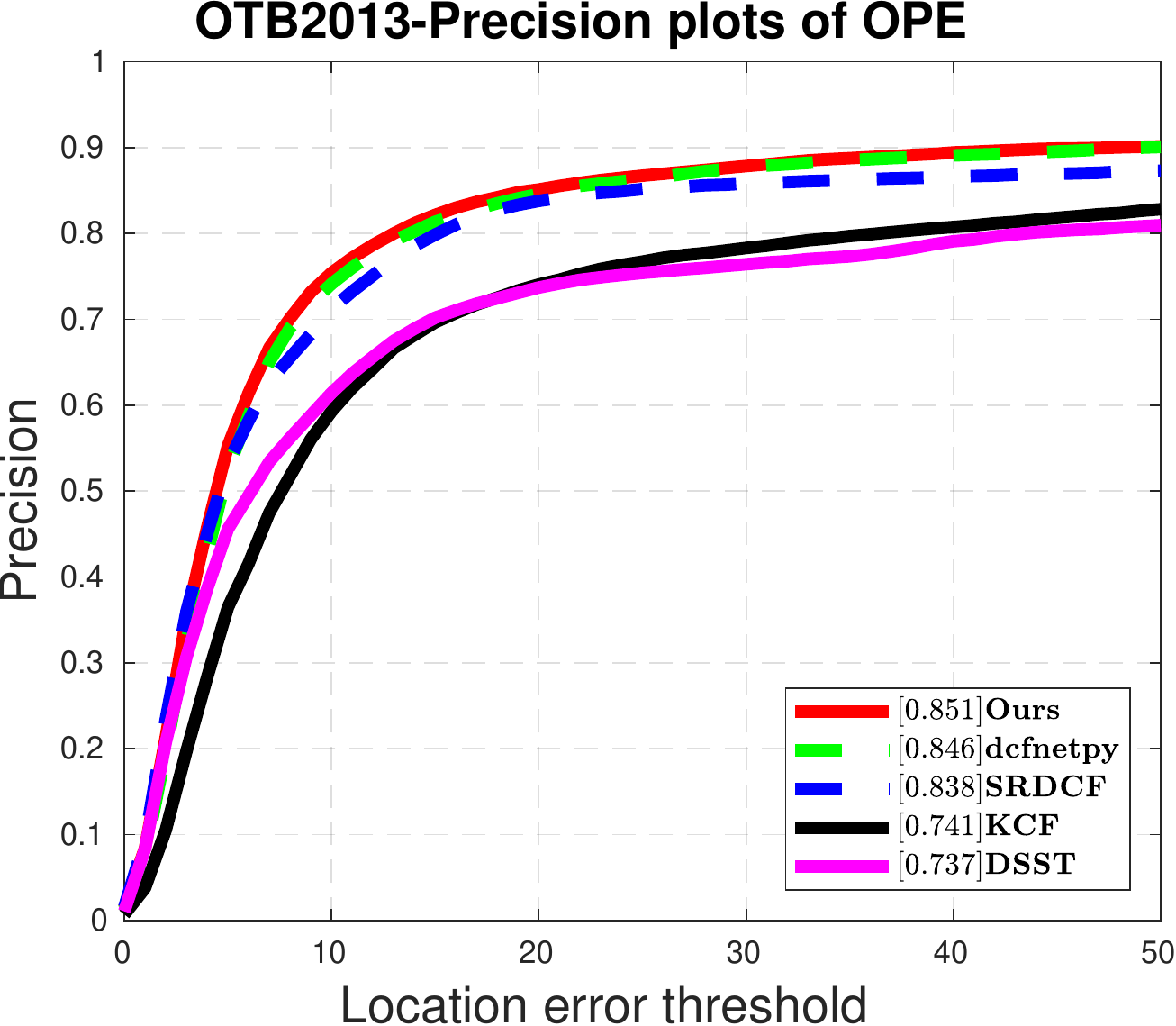}}
\subfigure[]{
\includegraphics[width=0.24\linewidth]{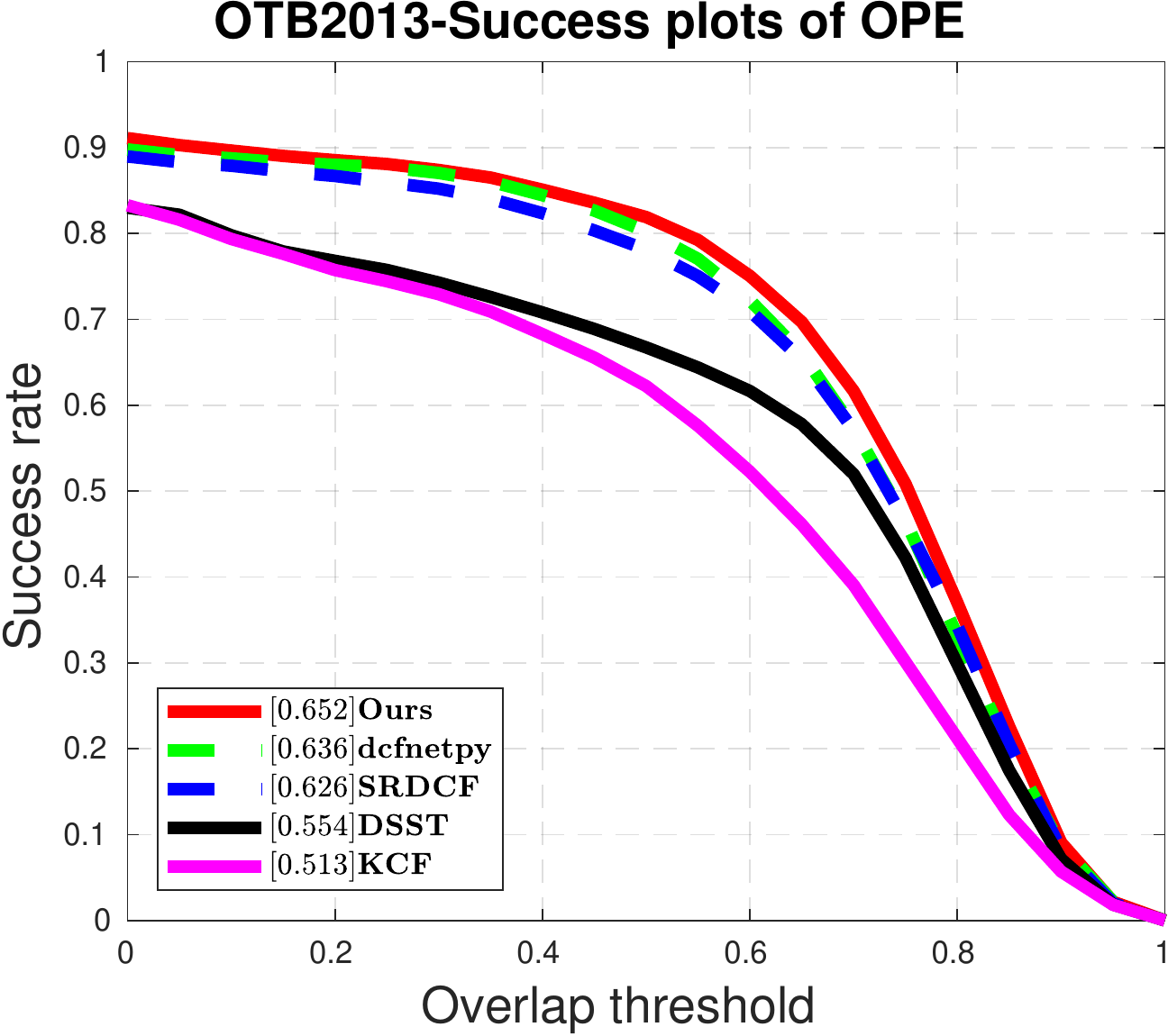}}
\subfigure[]{
\includegraphics[width=0.24\linewidth]{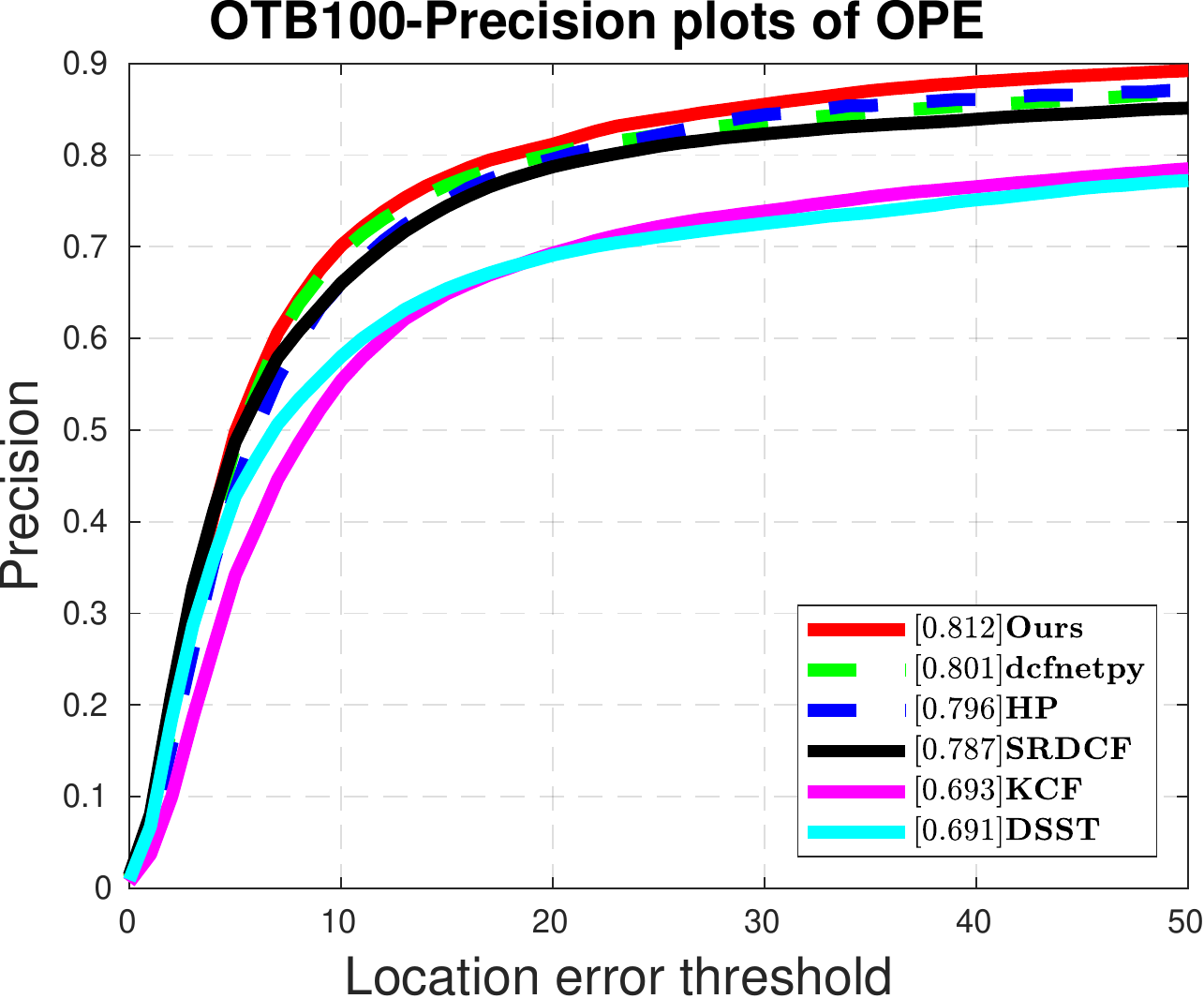}}
\subfigure[]{
\includegraphics[width=0.24\linewidth]{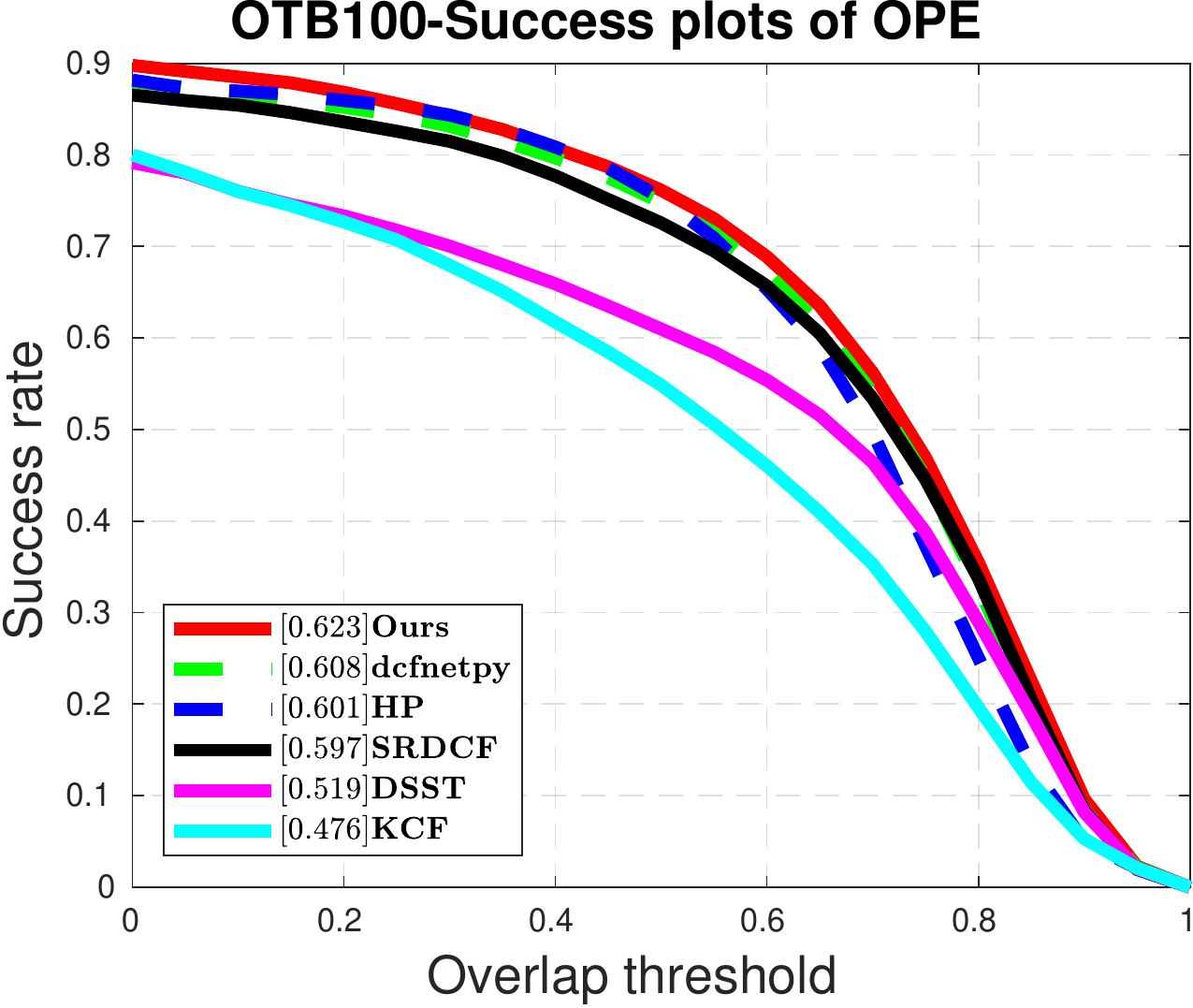}}

\caption{ Precision and success plots of overall performance comparison for the videos in the benchmark \cite{wu2013online}. Average distance precision and overlap success rate are reported. Listed CF based trackers are DSST~\cite{danelljan2014accurate}, KCF~\cite{henriques2015high}, SRDCF~\cite{danelljan2015learning}, dcfnet~\cite{wang2017dcfnet}, and HP~\cite{dong2018hyperparameter}.}
\label{overall}
\end{figure*}

\begin{figure*}[ht]
\centering
\subfigure[]{
\includegraphics[width=0.3\textwidth]{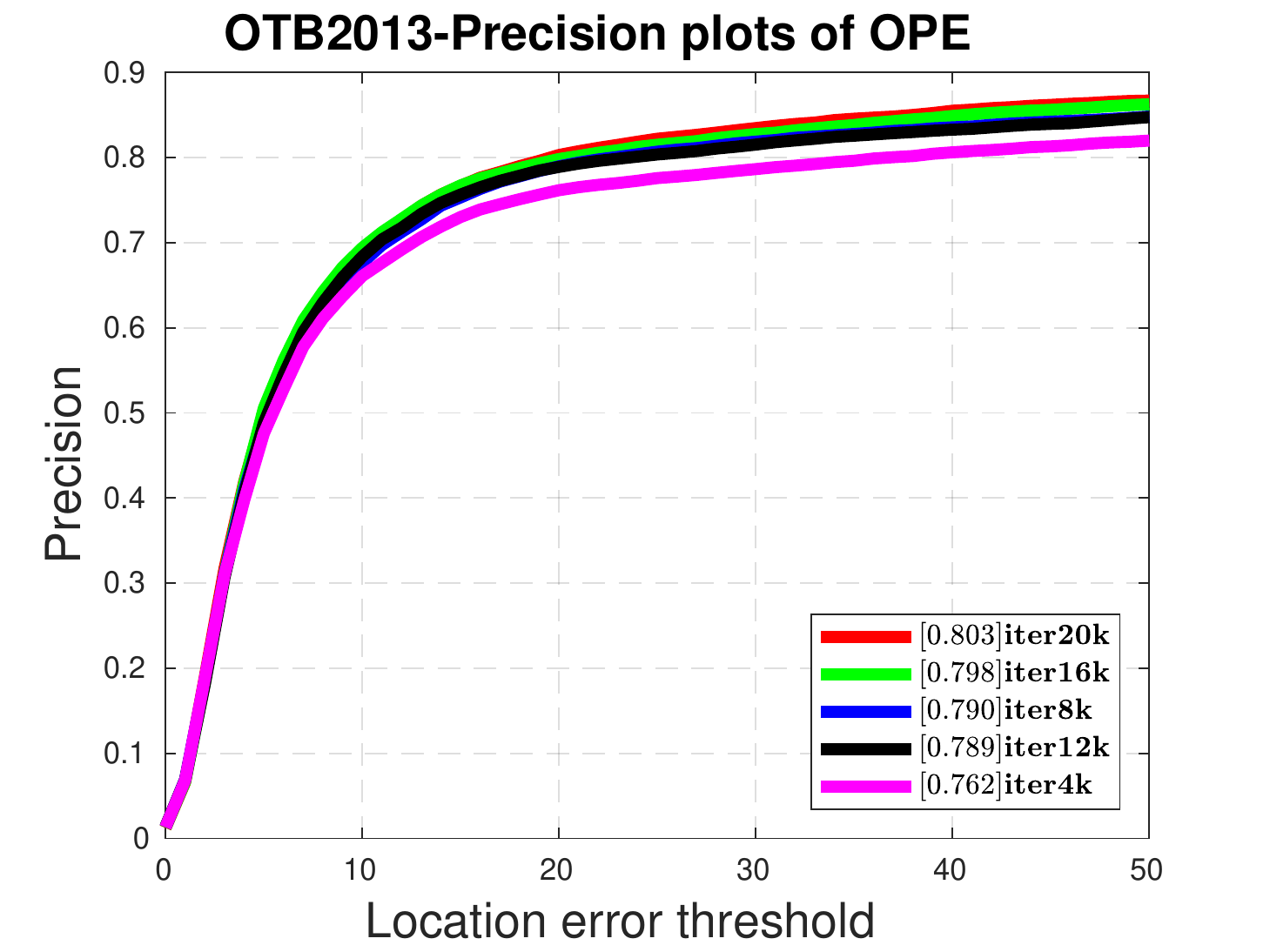}}
\subfigure[]{
\includegraphics[width=0.3\textwidth]{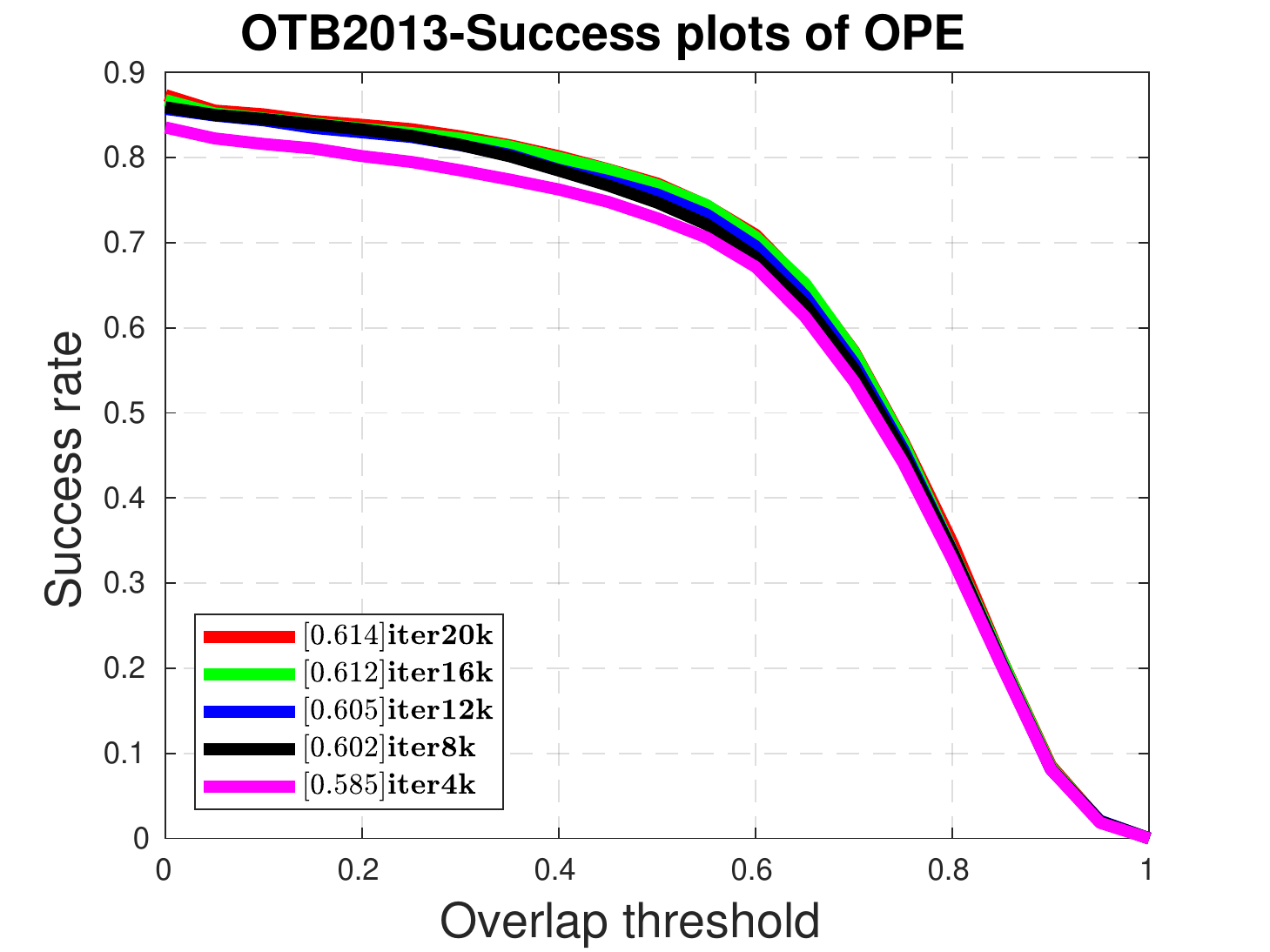}}

\caption{Tracking performance comparison with various reinforcement training iterations. Five different snapshots are shown, and the OPE performance increases with training iteration number.}
\label{fig_self}
\end{figure*}

\begin{figure*}[ht]
\centering
\subfigure[]{
\includegraphics[width=0.3\textwidth]{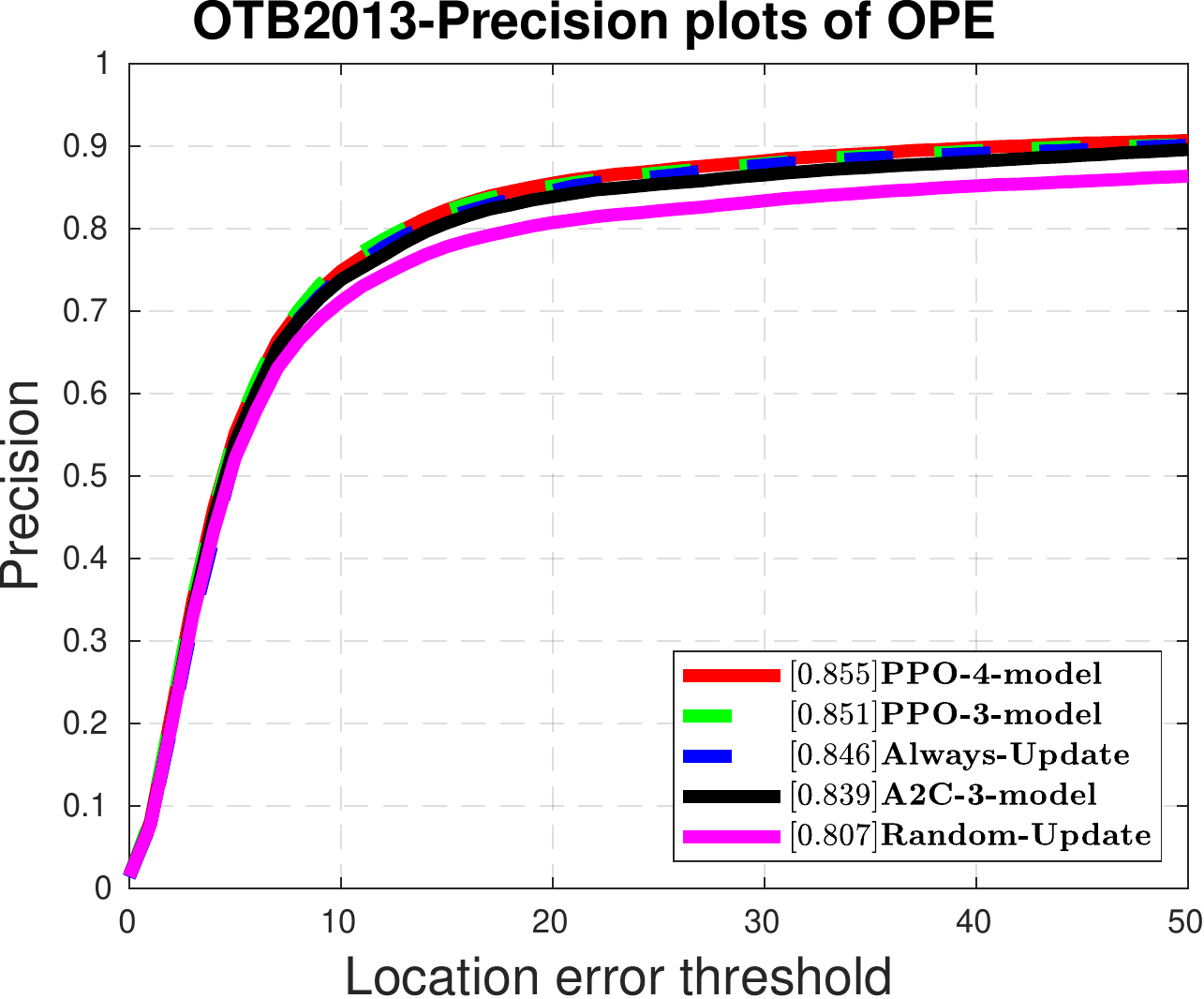}
}
\subfigure[]{
\includegraphics[width=0.3\textwidth]{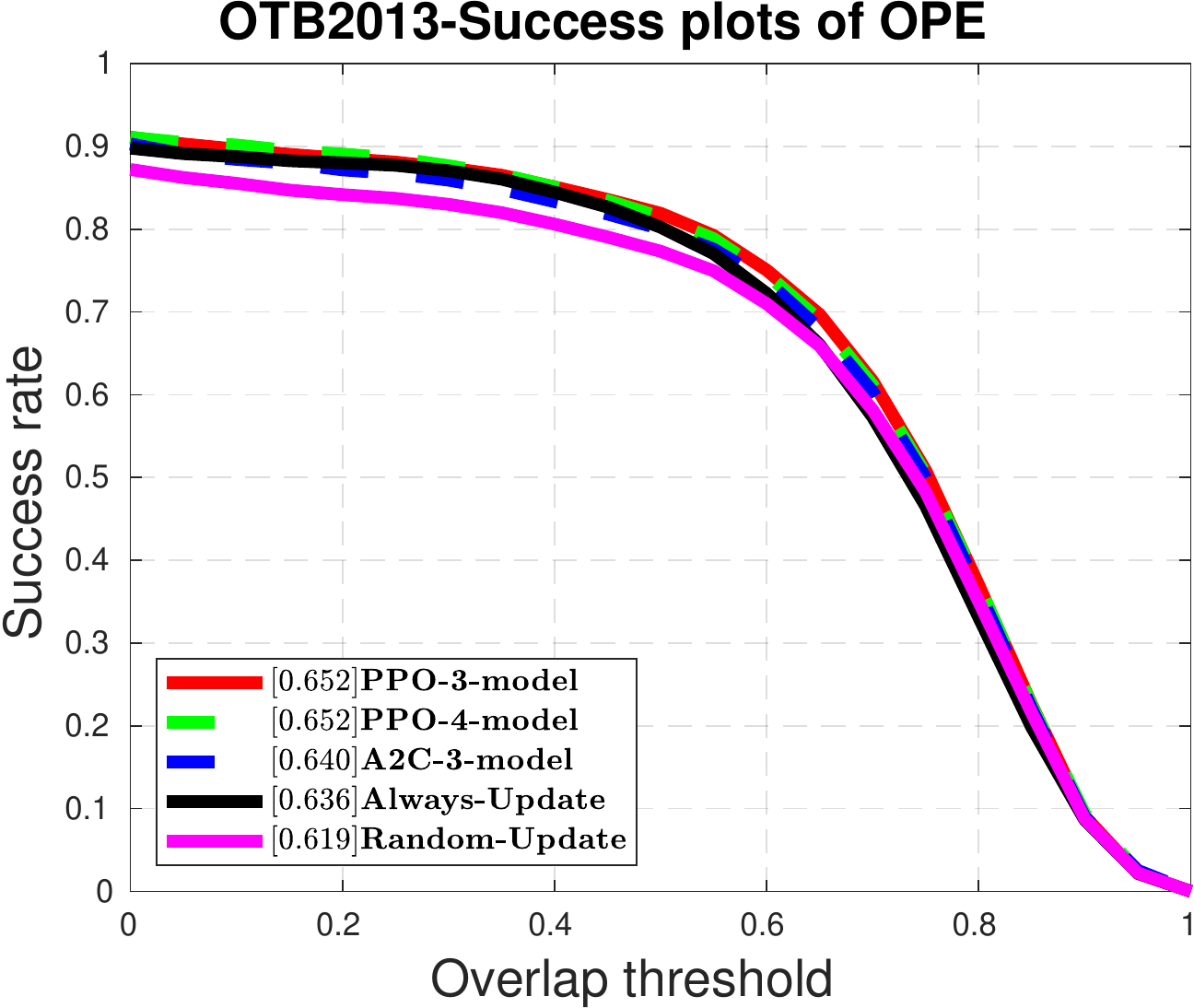} }

\caption{Tracking performance comparison of three different model update strategies: always updating CF model, random updating and the proposed updating by decision(PPO-3-model, PPO-4-model, A2C-3-model).}
\label{update_comp}
\end{figure*}

\textbf{Training Process} A training batch consists of randomly sampled sub-sequences and its ground-truth from the prepared database. It is noteworthy that the unexpected tracking failure would produce a series of useless negative samples, which means the length of training sequences should be limited. A simulation is used to generate a series of actions by the decision net, i.e., choosing one among different response maps from stored CF models.  Rewards will be obtained when the simulation is over, right actions with high expected returns will be encouraged with high rewards.
Finally, our decision net is trained to recognize appropriate CF models by optimizing the clipped surrogate objective function (\ref{equ_loss}).

Generally, in each iteration, our on-policy RL algorithm updates $\theta$ several times by gradient ascent, i.e.,  $\theta$ $\leftarrow$  $\theta$ $+$  $\Delta$ $\theta$. If the new policy $\pi$ or the new state value $V$ changes exceed a certain threshold, the clipped function will limit the network parameter update, which effectively constrains the variation caused by a challenging tracking sequence. This mechanism improves the training stability.

%\noindent
%\textbf{State and Action}

\begin{algorithm}
\caption{RL training via PPO}
\KwIn{\\Random sampled tracking sequence of length $L$, along with it ground-truth G\\
       Decision network $D({\theta})$}
\KwOut{Updated Decision network $D$}

\vspace{2ex}

Initialize CF model $M_0$ with the ground-truth \\
Set history CF model $M_1$, .. , $M_{K-1}$ = $M_0$ \\
\For{$t=1$ to $L$}
{
\For{$i=1$ to K}
{
Calculate response maps $P_i$ with each $M_i$\\
Produce confidence score via decision net $\pi(a_t|s_t;\theta)$ for each $P_i$
}
Choose the prediction map with the maximum confidence;\\
Localize the target according to chosen $P_m$;\\
Obtain reward $r_m$\\
Update corresponding CF models and save to history;

}
Sum discounted reward as return\\
Update Decision network $D({\theta})$ by Adam with equation (\ref{equ_loss}) for $d=10$ times\\
\end{algorithm}

\section{Evaluation}
\label{sec4}

In this section, we detail our experimental setup and the parameters we used during the training and testing. Quantitative and qualitative experiments have been conducted on popular visual tracking benchmark datasets, namely the OTB2013 and OTB100. We compared our proposed algorithm with the other five CF-based tracking frameworks. Meanwhile, we validated the
effectiveness of our proposed method by conducting various ablation studies.  For fair comparisons, no additional modification is allowed during the evaluation. The experiments were conducted on a system with an E5-2620v3 2.4GHz CPU having 32~GB memory and a GTX TITAN X GPU using MATLAB2017b and PyTorch.

\subsection{Experimental Setup}
In the tracking process, the searching image within the current frame is twice the target size on the horizontal and vertical direction. In order to cover different scale changes, 3 scaled versions of the search image are used to find the best scale that fits the scale change.
The scale parameter is set to $1.025$. If not explicitly specified,three CF models are maintained in our experiments, i.e., $k=3$, including the initial model for tracking, the dynamic model, and the accumulated model. The accumulated model is updated at each frame, while the initial model is kept unchanged. The dynamic model is updated once it is selected by the decision network. For the dynamic model and accumulated model, average moving parameter $0.05$ is used, and new CF models will replace old models.

\subsection{Comparison on Benchmarks}
We evaluated our method in comparison with existing CF trackers on the popular visual tracking benchmarks, Object Tracking Benchmark (OTB)~\cite{wu2013online}. Tracking algorithms KCF ~\cite{henriques2015high}, DSST~\cite{danelljan2014accurate}, SRDCF~\cite{danelljan2015learning}, DCFnet~\cite{wang2017dcfnet} and HP~\cite{dong2018hyperparameter} are evaluated for comparison. The \textit{dcfnetpy} is our implemented algorithm of DCFnet~\cite{wang2017dcfnet} in python, which achieved similar performance as reported in~\cite{wang2017dcfnet}.
Two standard evaluation metrics, namely distance precision (DP) and overlap success (OS) rate, are used to evaluate trackers' performance. DP is the frame proportion of the predicted position within a given threshold. Overlap success rate is defined as the percentage of frames that overlap between predicted location and ground-truth surpassing the threshold.

\begin{table}[h]
\begin{center}
\caption{A comparison of our approach with other CF-based trackers. The mean overlap precision (OS) (\%) and distance precision (DP) (\%) over all the videos in the OTB2013 dataset are presented. DP at a threshold of 20 pixels, overlap success (OS) rate at an overlap threshold 0.6.}
\scriptsize
\begin{tabular}{|c|c|c|c|c|c|c|}
\hline
Method & Proposed &  \textit{dcfnetpy}   &  SRDCF~\cite{danelljan2015learning}  & DSST~\cite{danelljan2014accurate}  & KCF~\cite{henriques2015high}   \\
%& (VGGNet) & (ResNet-50) & (ResNet-101) & (ResNet-50) & (ResNet-50) & (ResNet-101) \\
\hline\hline
OS (\%) & \textbf{74.58} & $72.23$ & $70.98$  & $61.65$ & $52.25$ \\%& \textbf{77.23} \\
DP (\%) & \textbf{85.12} & $84.59$ & $83.79$  & $73.70$ & $74.06$ \\%& \textbf{80.45} \\
\hline
\end{tabular}
\label{tab2}
\end{center}
\end{table}

\begin{table}[h]
\begin{center}
\caption{A comparison of our approach with other CF-based trackers. The mean overlap precision (OS) (\%) and distance precision (DP) (\%) over all the 100 videos in the OTB100 dataset are presented. DP at a threshold of 20 pixels, overlap success (OS) rate at an overlap threshold 0.6.}
\scriptsize

\begin{tabular}{|c|c|c|c|c|c|c|}
\hline
Method & Proposed &  \textit{dcfnetpy}   &  SRDCF~\cite{danelljan2015learning}  & DSST~\cite{danelljan2014accurate}  & KCF~\cite{henriques2015high}   \\
%& (VGGNet) & (ResNet-50) & (ResNet-101) & (ResNet-50) & (ResNet-50) & (ResNet-101) \\
\hline\hline
OS (\%) & \textbf{68.89} & $67.07$ & $65.67$  & $55.39$ & $46.03$ \\%& \textbf{77.23} \\
DP (\%) & \textbf{81.19} & $80.13$ & $78.74$  & $69.10$ & $69.31$ \\%& \textbf{80.45} \\
\hline
\end{tabular}
\label{tab3}
\end{center}

\end{table}

%####################################################################################################  Mod
\begin{table}[h]
\begin{center}
\caption{Tracking performance comparison with various reinforcement training iterations. On OTB2013, DP at a threshold of 20 pixels, overlap success (OS) rate at an overlap threshold 0.6.}
\scriptsize

\begin{tabular}{|c|c|c|c|c|c|c|}
\hline
Iteration & \textit{4k} &  \textit{8k}   &  \textit{12k}  & \textit{16k}  & \textit{20k}   \\
%& (VGGNet) & (ResNet-50) & (ResNet-101) & (ResNet-50) & (ResNet-50) & (ResNet-101) \\
\hline\hline
OS (\%) & 67.2 & $68.3$ & $69.6$  & $70.6$ & $70.9$ \\%& \textbf{77.23} \\
DP (\%) & 76.2 & $78.9$ & $79.0$  & $79.8$ & $80.3$ \\%& \textbf{80.45} \\
\hline
\end{tabular}
\label{tab3}
\end{center}

\end{table}

\begin{table*}[h]
\begin{center}
\caption{Tracking performance comparison of 5 different model update strategies and test On OTB2013, DP at a threshold of 20 pixels, overlap success (OS) rate at an overlap threshold 0.6.}
\scriptsize

\begin{tabular}{|c|c|c|c|c|c|c|}
\hline
Method & PPO-4-model & PPO-3-model &  A2C-3-model    & Always-Update   &  Random-Update    \\

\hline\hline

OS (\%) & \textbf{75.05} & \textbf{74.58} & $73.74$ & $72.33$ & $71.00$ \\%& \textbf{80.45} \\
DP (\%) & \textbf{85.48} & \textbf{85.12} & $83.85$ & $84.59$ & $80.72$ \\%& \textbf{77.23} \\
\hline
\end{tabular}
\label{tab3}
\end{center}

\end{table*}

%####################################################################################################  End

\noindent
\textbf{Quantitative Comparison}
Overall performance comparison for the 51 videos in the benchmark \cite{wu2013online} is reported in Fig.~\ref{overall}, which includes both
precision and success plots.
It can be observed that in success plots our proposed algorithm is always above other trackers for overlap threshold above 0.5.
The performance gain is increasing with overlap threshold,
showing our proposed method consistently contributes to the tracking accuracy with various overlap threshold.

Table~\ref{tab2} is comparisons of our approach with other CF-based trackers.
The mean overlap precision (OS) and distance precision (DP) over all the OTB dataset are presented.
The results are obtained with DP at a threshold of 20 pixels, overlap success(OS) rate at an overlap threshold of 0.6.
Results show that our algorithm performs favorably against other CF methods for a common setting.
Among the existing CF trackers, our proposed method achieves the best results with an OS of 68.89\%, DP of 81.19\% on OTB100.
Our achieved OS and DP are respectively 1.82\%  and 1.06\% higher than that of CF models without model selection (\textit{dcfnetpy}).

In Fig.~\ref{attr}, the performance of 5 CF-based trackers for 11 attributes on OTB100 is reported, including background clutter, low resolution, scale variation, illumination variation, deformation, motion blur, in-plane rotation, occlusion, and out-of-view. Generally, our proposed tracker achieves superior accuracy compared to other CF trackers for most of the attributes. Due to the multiple model selection, our method is able to handle occlusion better during the tracking, and the results in 48 occlusion sequences improve by 3.7\% in success rate and 3.1\% in precision compared with the always updating strategy (\textit{dcfnetpy}). Similarly, our proposed method works well in 14 out of view sequences and in 9 low-resolution sequences.

\noindent
\textbf{Qualitative Comparison} Fig.~\ref{fig_vis} presents the superiority of our algorithm qualitatively compared to other 4 CF trackers on 7 challenging sequences.
The CF2, DSST methods lose track of the target gradually due to significant occlusion and motion blur in Box and Girl sequences.
The SRDCF, KCF, CF2 trackers are not able to keep tracking the target after occlusion and illumination changes in Box and Girl2 sequences. It can also be observed that when scale variation and occlusion happen as in Dragonbaby, the DSST and KCF trackers do not perform well. Other trackers fail in the presence of out-of-plane rotation, scale variation, and fast motion.
It is noticed that our proposed multiple model selection could discover the missing target after a long-term tracking failure, while
other trackers can hardly recover from the drifting.
Overall, our proposed tracker is able to alleviate the drifting issue in many challenging sequences.

\subsection{Ablation Study}

We conducted some ablation studies to demonstrate the effectiveness of our method.
In Fig.~\ref{fig_self}, performance is reported with different training iterations on OTB2013.
The precision and success rates increase with the iteration, proving that the reinforcement learning process effectively guides the optimization.

\begin{figure}[ht]
\centering

\includegraphics[width=0.4\textwidth]{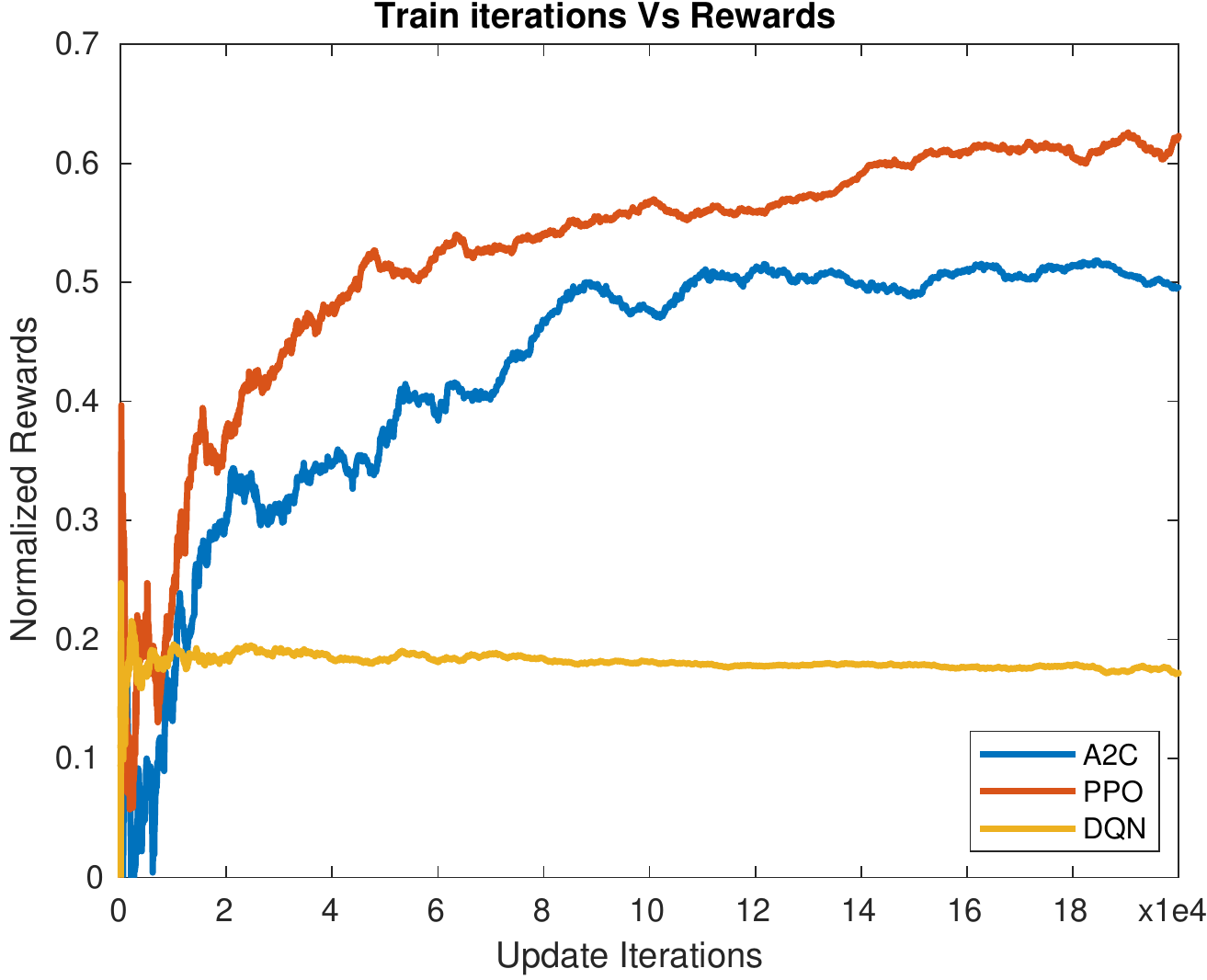}

\caption{Normalized Rewards vs Iteration Number through train process, A2C~\cite{mnih2016asynchronous}, PPO~\cite{schulman2017proximal}, and DQN~\cite{mnih2015human}}
\label{rewards}
\end{figure}

%####################################################################################################  xyc
We also conducted additional experiments with different CF model selection/updating schemes.
The "always-update" scheme always uses the latest CF model to find the tracking object and updates the model at each frame.
The "random-update" scheme randomly select a model among the initial model, dynamic model and accumulated model, and
updates the randomly selected model at each frame. We set dynamic model with number 1 and 2 respectively in our experiments, which are denoted by PPO-3-model and PPO-4-model.
The performance comparison results are plotted in Fig.~\ref{update_comp} and numeric results are reported in Table ~\ref{tab3}. 
Overall, our model selection scheme outperforms both the "always-update" and "random-update" schemes, showing that by using the proposed CF selection strategy, our decision network is able to choose the most suitable CF for visual tracking, and to a certain extent, the model drift has been reduced. PPO-model-3 and PPO-model-4 lead to similar performance for both metric OS and DP. Therefore, we use PPO-model-3 model throughout the paper for performance evaluation due to its lower complexity.

We also employ different reinforcement learning algorithms to replace the proximal policy optimization. First, we disable the Clipped Surrogate term and degrade it into a basic synchronous advantage actor-critic model (A2C~\cite{mnih2016asynchronous}). The policy/value network structures and parameters are kept the same. The update is performed after the 4 actor-learners finishing collecting data, in order to improve the training stability. Moreover, we further degrade the RL algorithm into a DQN~\cite{mnih2015human}. The agent's experiences state at each timestep is stored to perform experience replay, Q-learning updates are applied by random sampling from the experience pool.
The train reward with update iterations is shown in Fig.~\ref{rewards}. It shows that an improvement of the reward due to the advantage actor-critic algorithm A2C and PPO, while the traditional DQN does not work for the model selection under a similar training setting. It is also important to note that the A2C algorithm takes $50\%$ more time to reach the same update iteration number of PPO in our experiments. Also the PPO algorithm ends up with higher rewards than the A2C algorithm and a better tracking performance is achieved,as reported in Table~\ref{tab3}. 
 
%####################################################################################################  End
\section{Conclusions}
\label{sec5}

In this paper, we have proposed a novel approach for CF-based visual tracking. In our approach, multiple CF models are updated and maintained in parallel and an optimal model is selected on demand using the deep reinforcement learning.
The proposed algorithm learns the model selection policy with the proximal policy optimization algorithm,
while utilizing the selected CF model to conduct object tracking.

We show that the model selection via response map can effectively overcome the model drifting issues, and enhance
the robustness of the trackers. Our exhaustive experimental evaluation using two key benchmarks, covering both the quantitative and qualitative aspects, show that our approach can handle a number of tracking challenges and can offer substantially better tracking performance when compared to traditional CF-based trackers.

% For future study, it is noticed that the CF model updating rate should be adjustable, which is worthy to be further studied and integrated into the reinforcement learning mechanism.

\section{aknowledgement}
The authors would like to thank Prof.Tammam Tillo for his valuable contributions to this work.

\ifCLASSOPTIONcaptionsoff
  \newpage
\fi

\begin{figure*}[ht]
\centering
\includegraphics[width=0.95\textwidth]{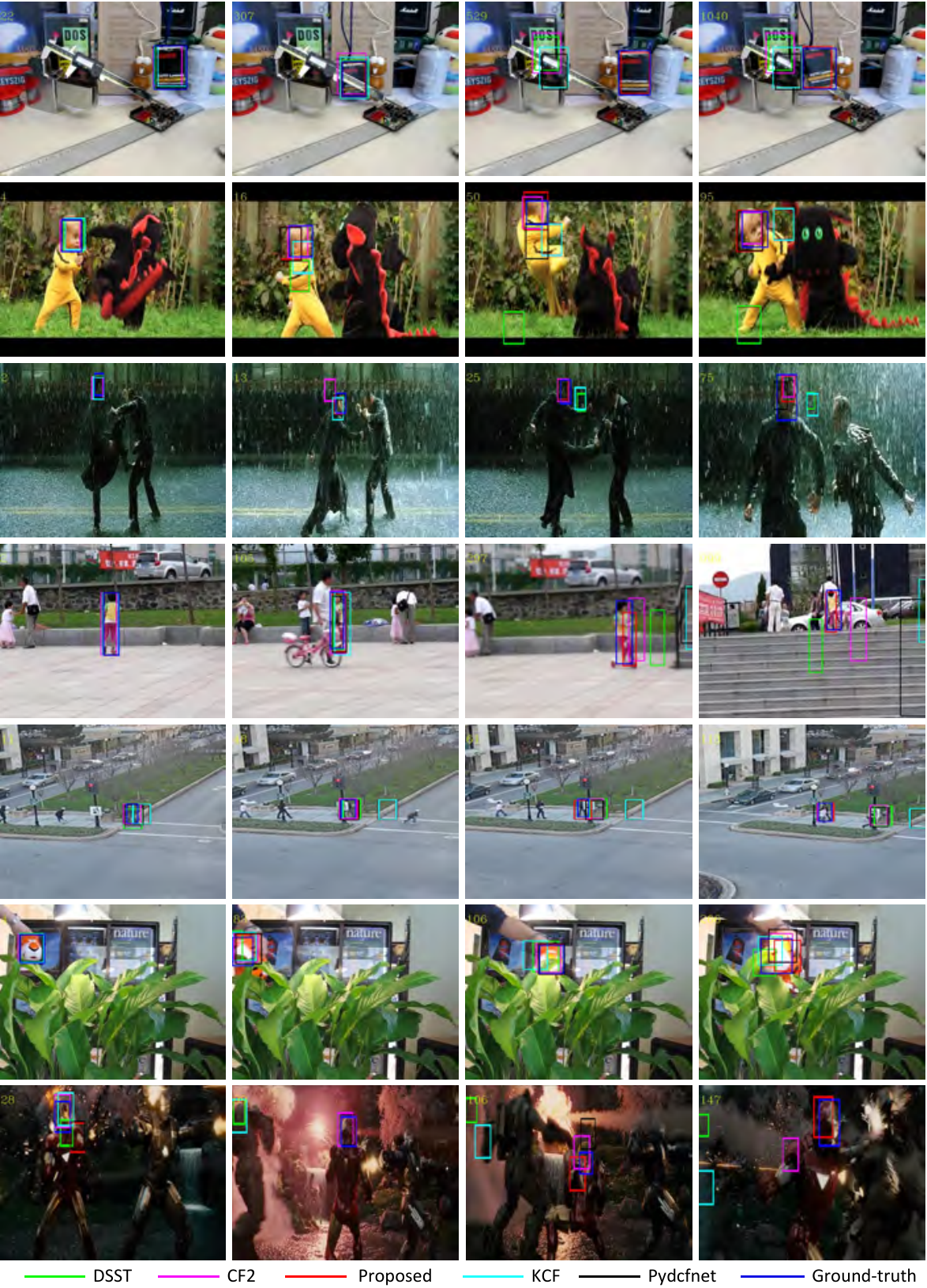}

\caption{Visualizations of our tracking results(Box, DragonBaby, Matrix, Girl2, Human, Tiger, Ironman). Green, Purple, Red, Light Blue, and Black box denote tracking results of DSST, CF2, Proposed, KCF, \textit{pydcfnet}, respectively. Blue box is the ground-truth box, Yellow numbers on the top-left corners indicate frame numbers. }
\label{fig_vis}
\end{figure*}

\begin{figure*}[ht]
%\centering
\includegraphics[width=0.23\textwidth]{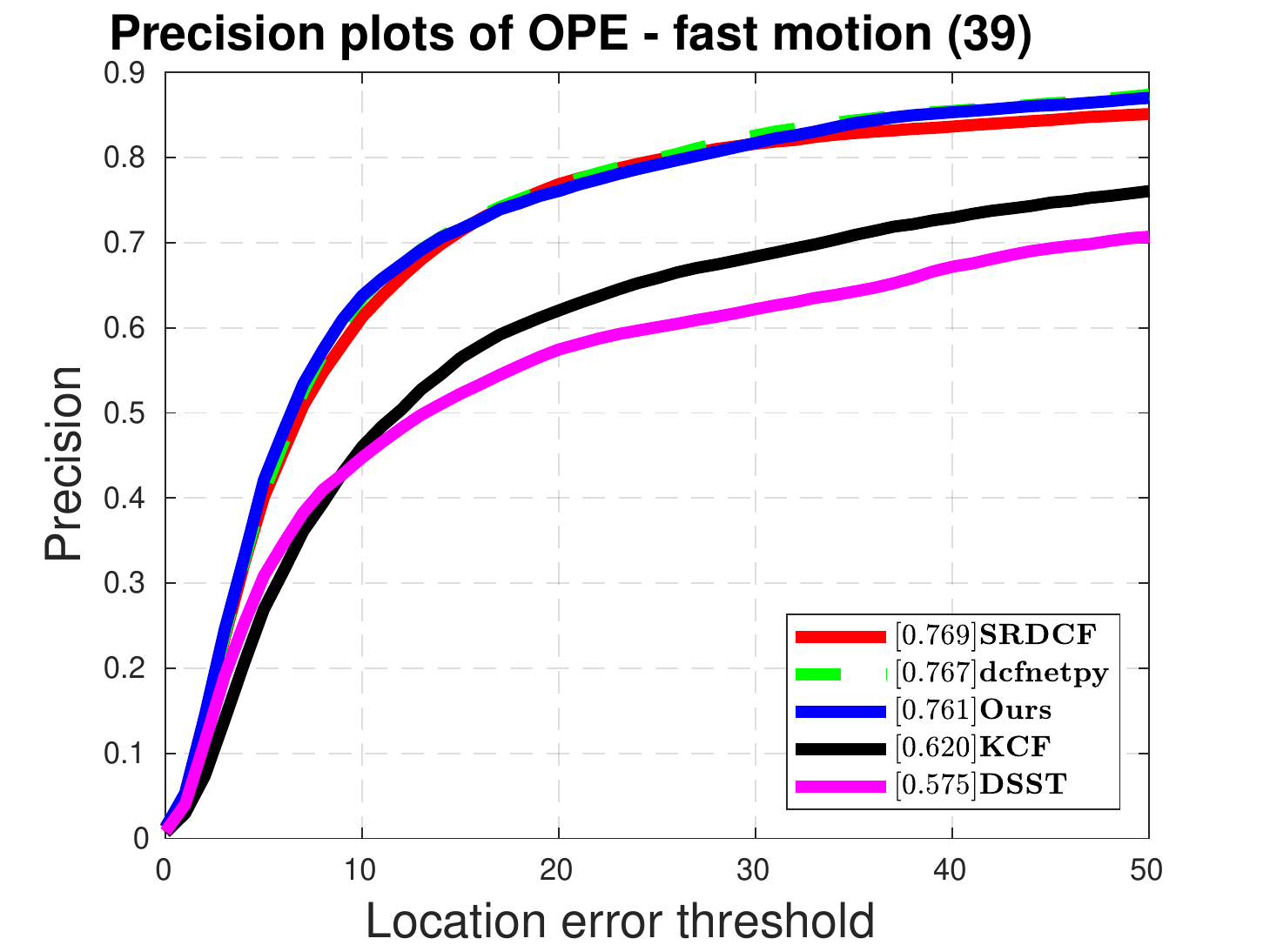}
\includegraphics[width=0.23\textwidth]{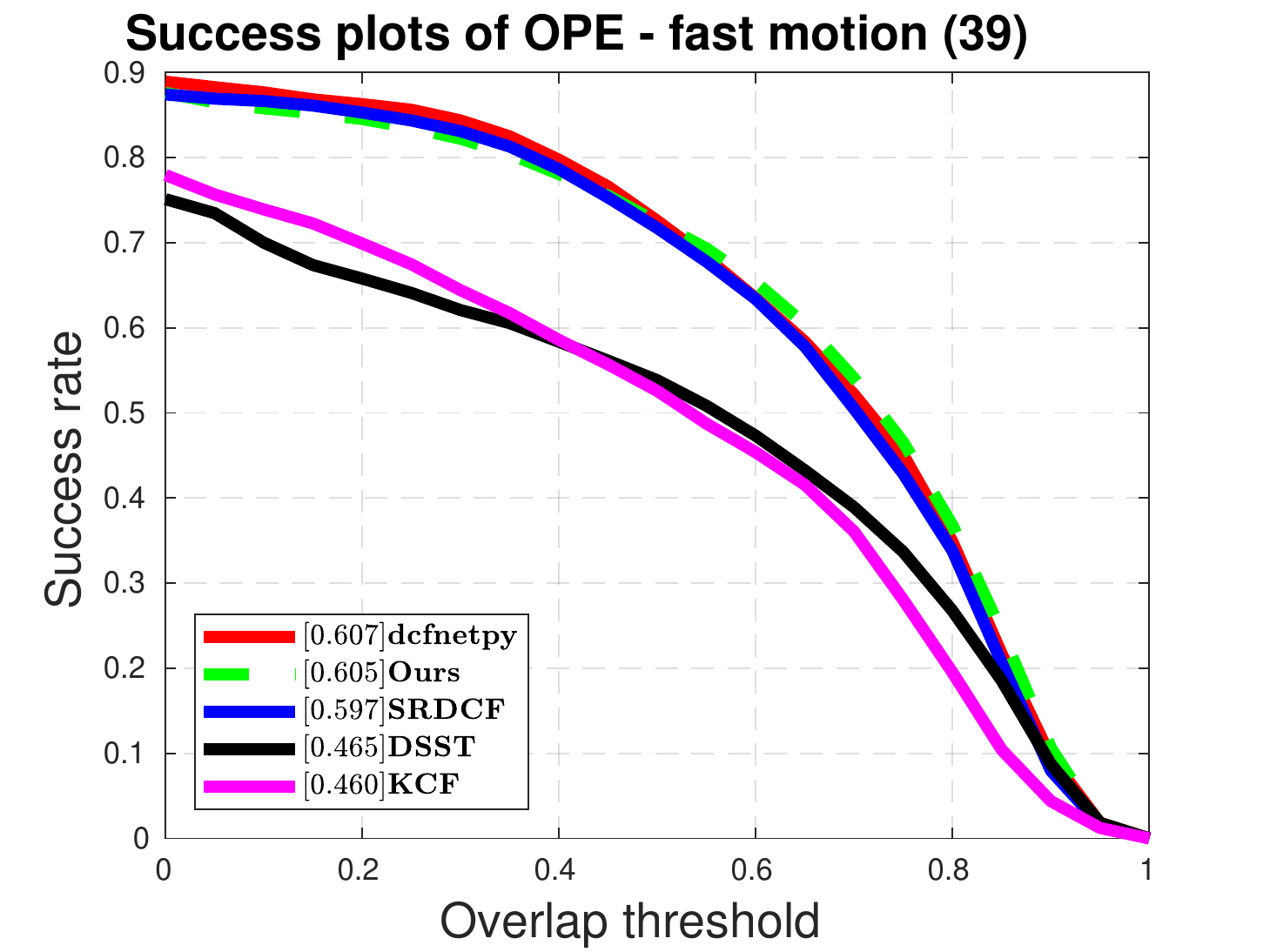}
\includegraphics[width=0.23\textwidth]{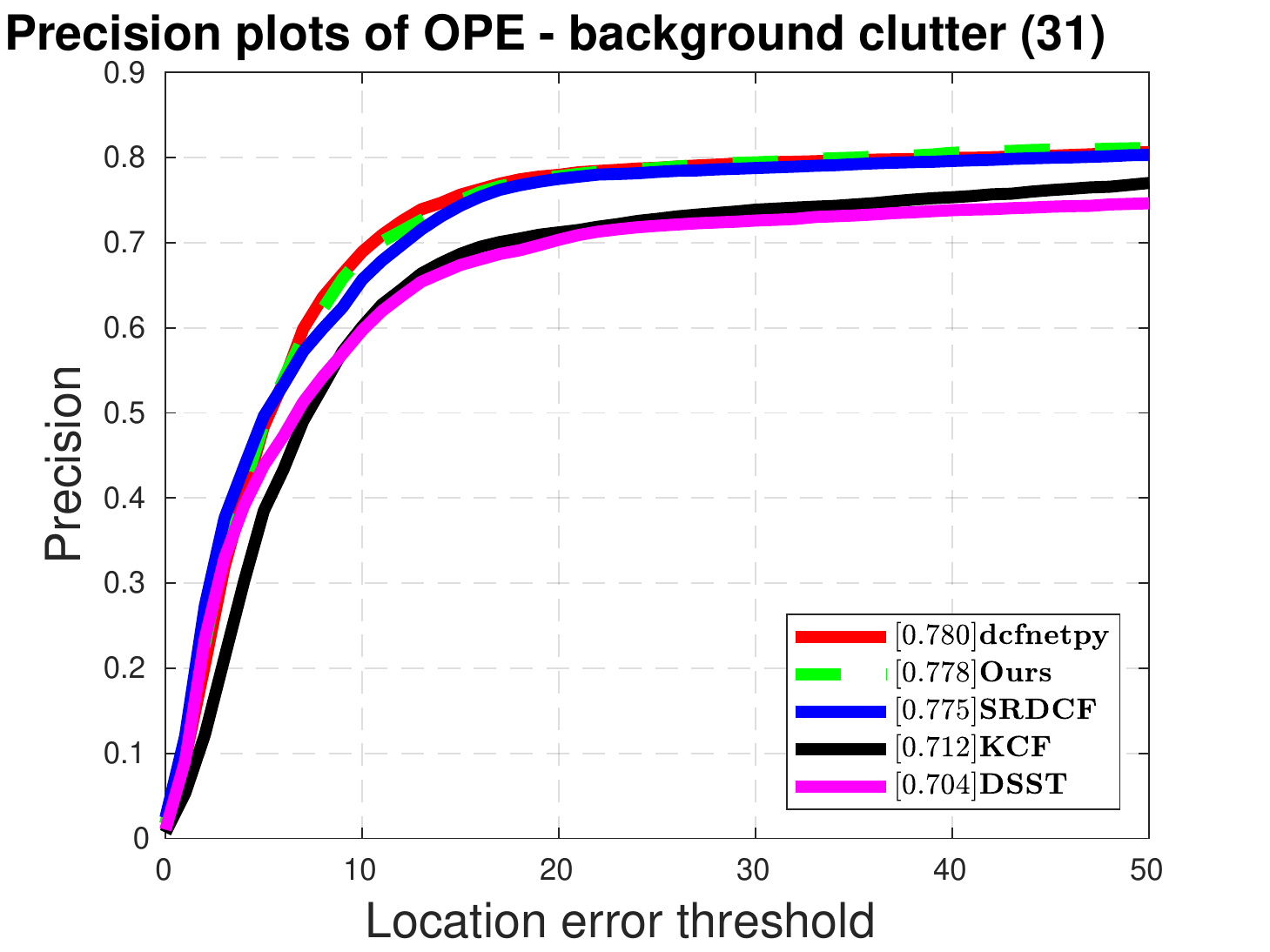}
\includegraphics[width=0.23\textwidth]{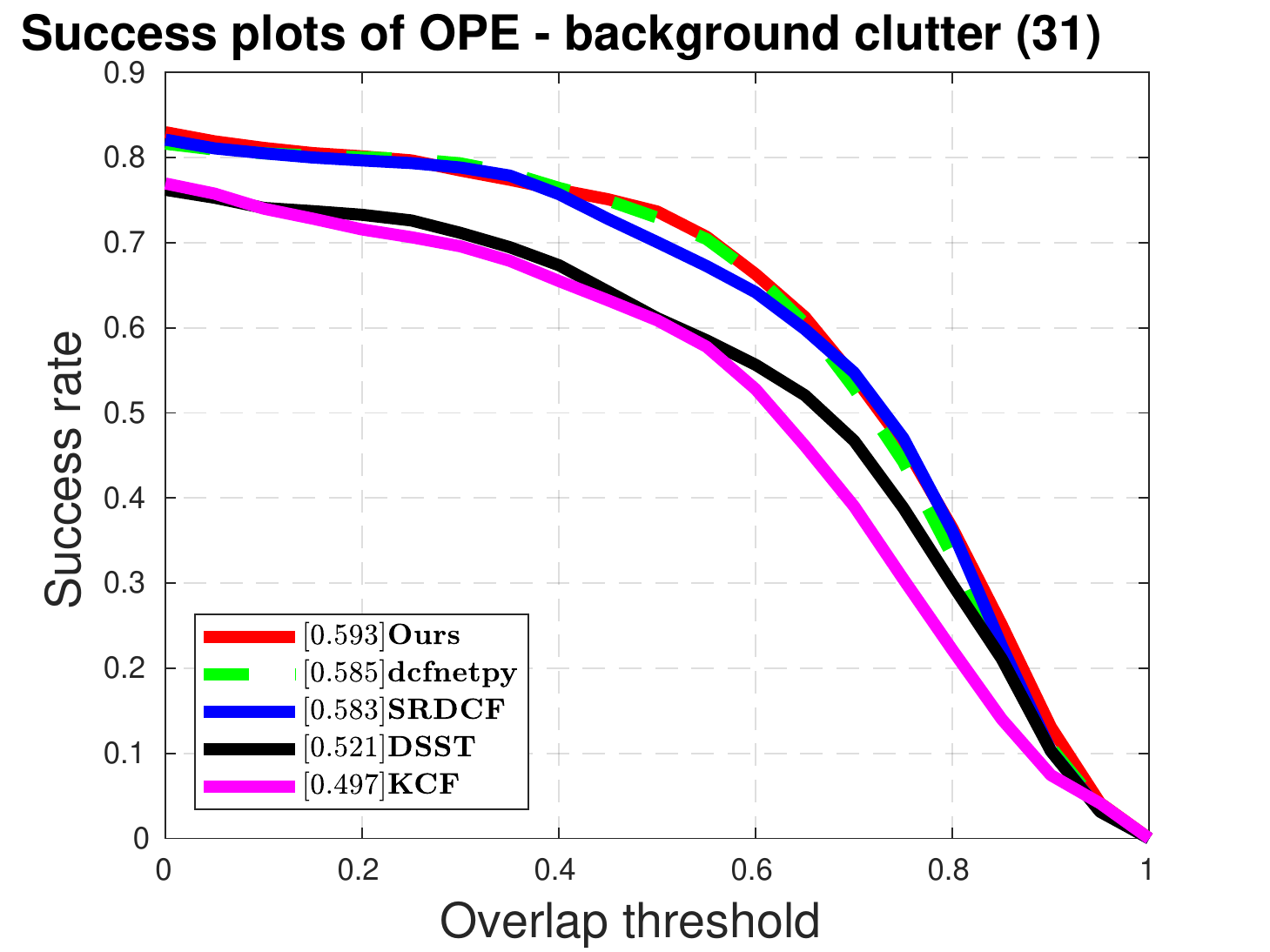}\\
\includegraphics[width=0.23\textwidth]{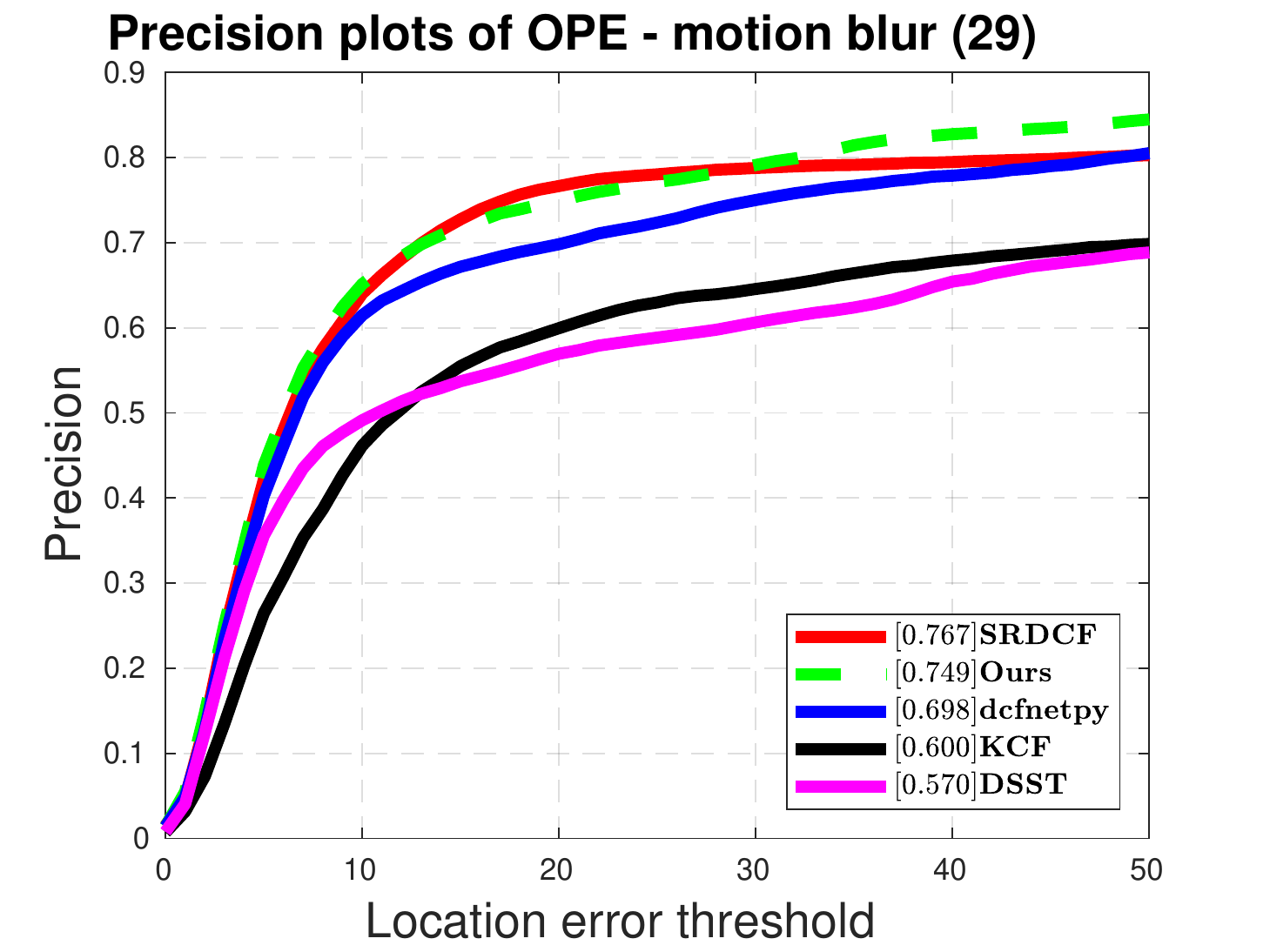}
\includegraphics[width=0.23\textwidth]{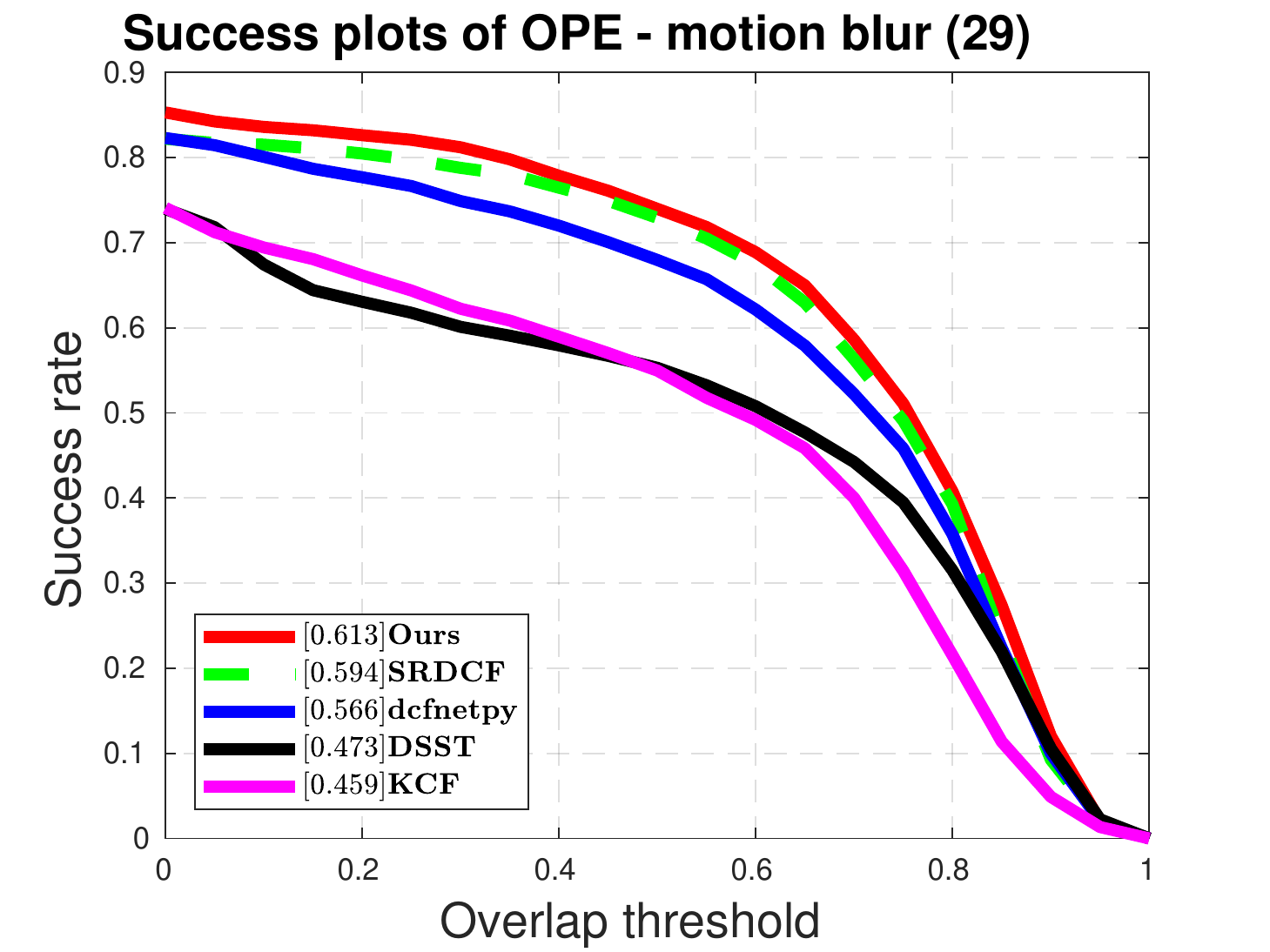}
\includegraphics[width=0.23\textwidth]{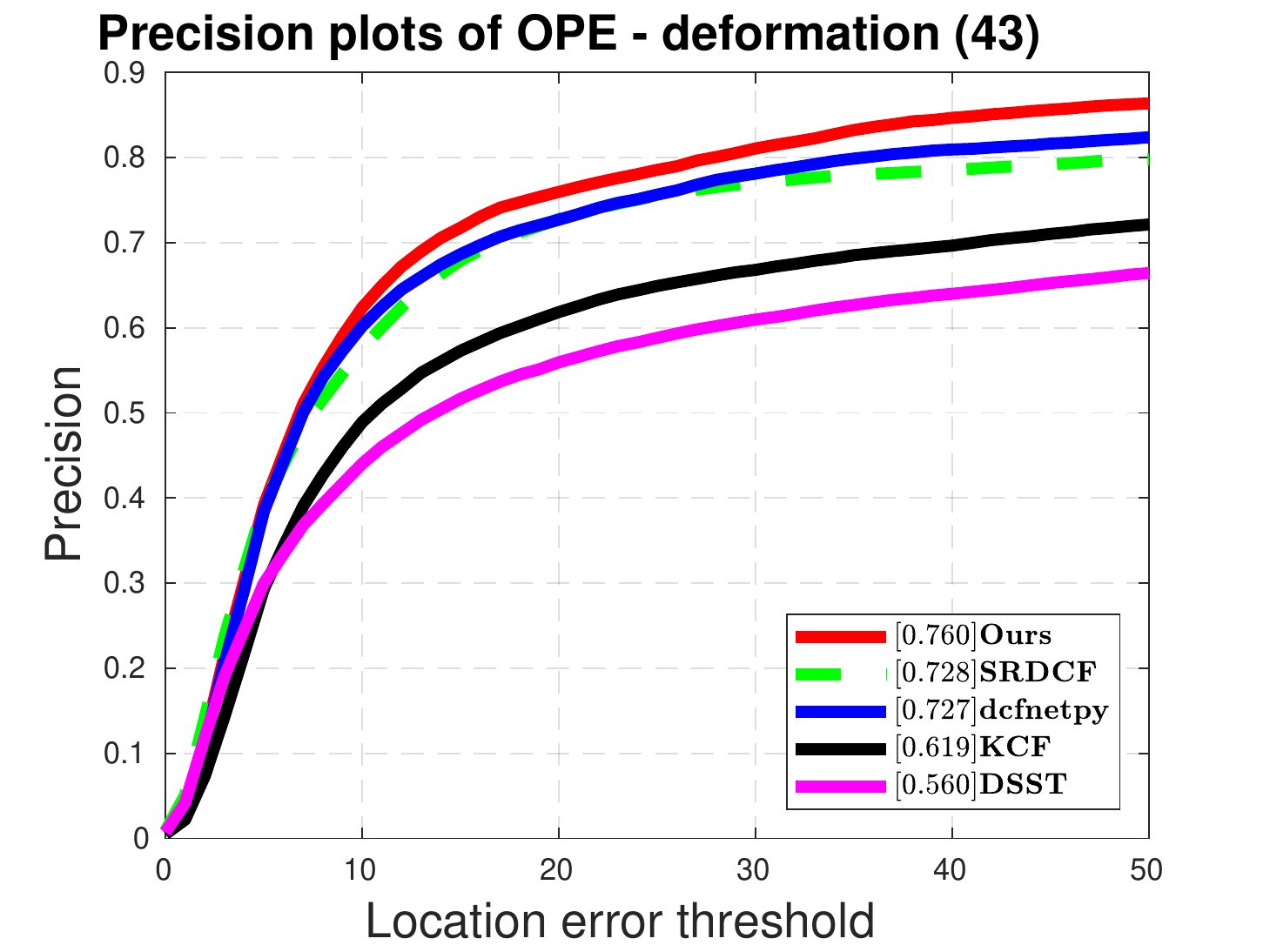}
\includegraphics[width=0.23\textwidth]{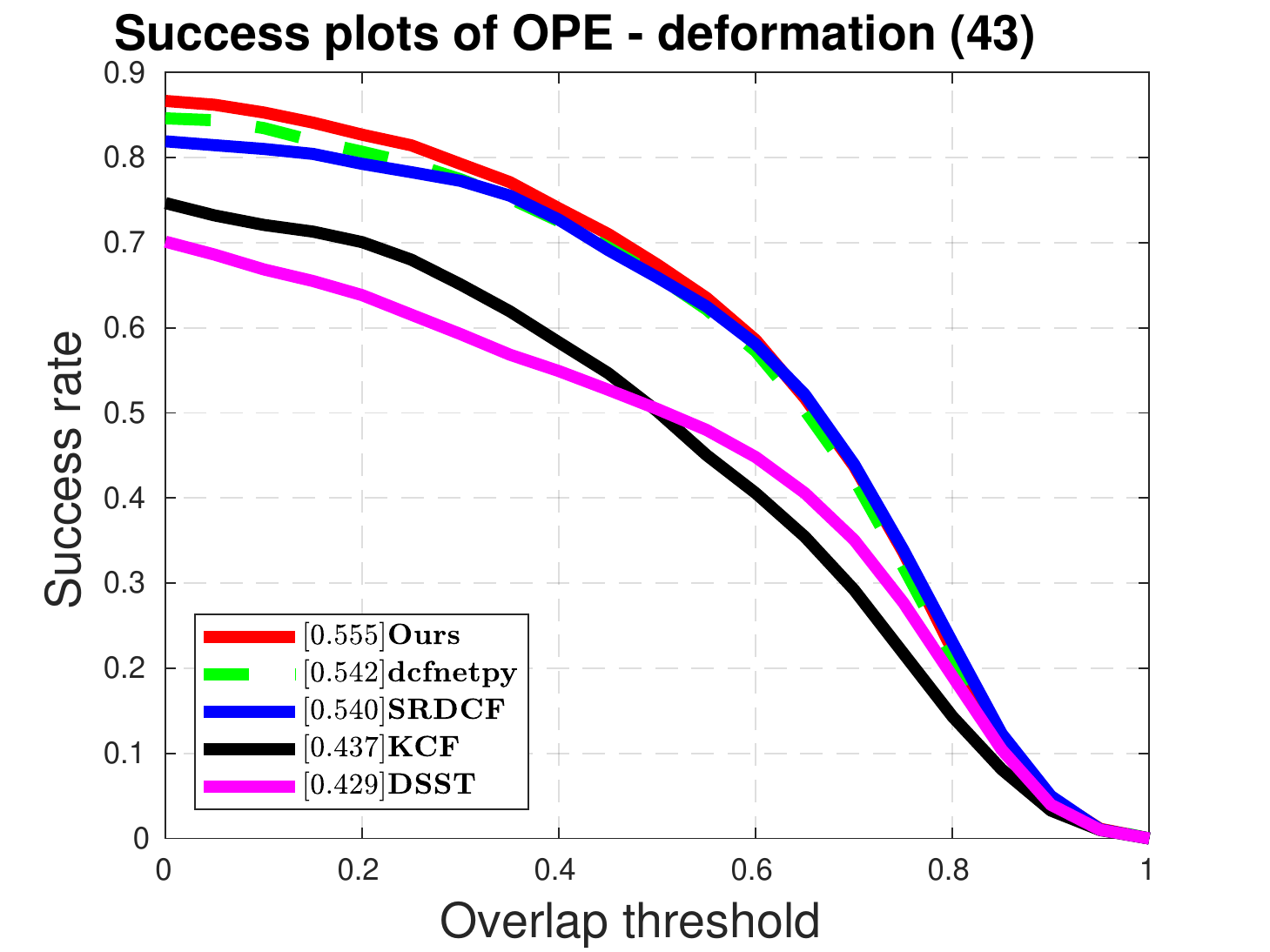}\\

\includegraphics[width=0.23\textwidth]{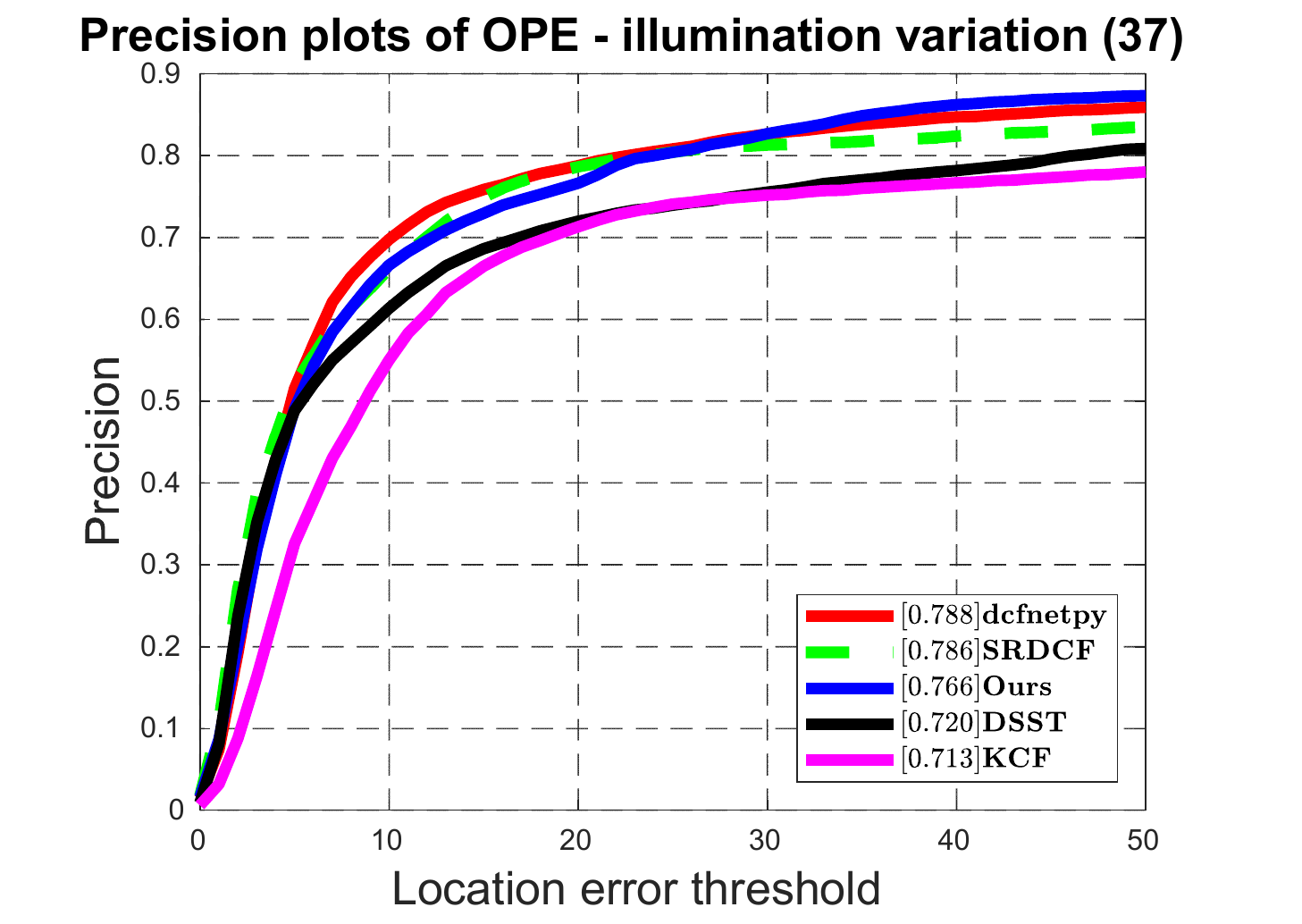}
\includegraphics[width=0.23\textwidth]{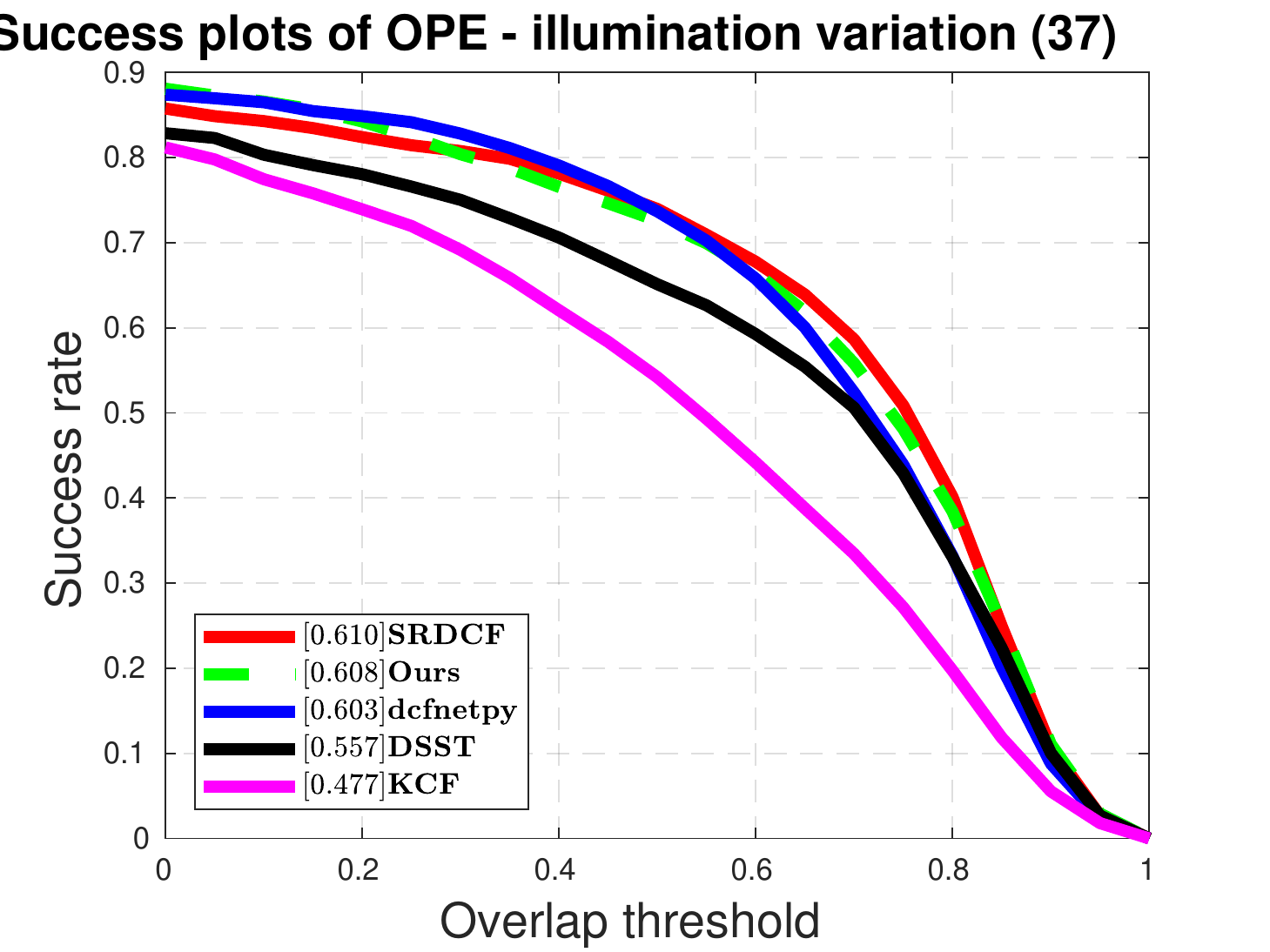}
\includegraphics[width=0.23\textwidth]{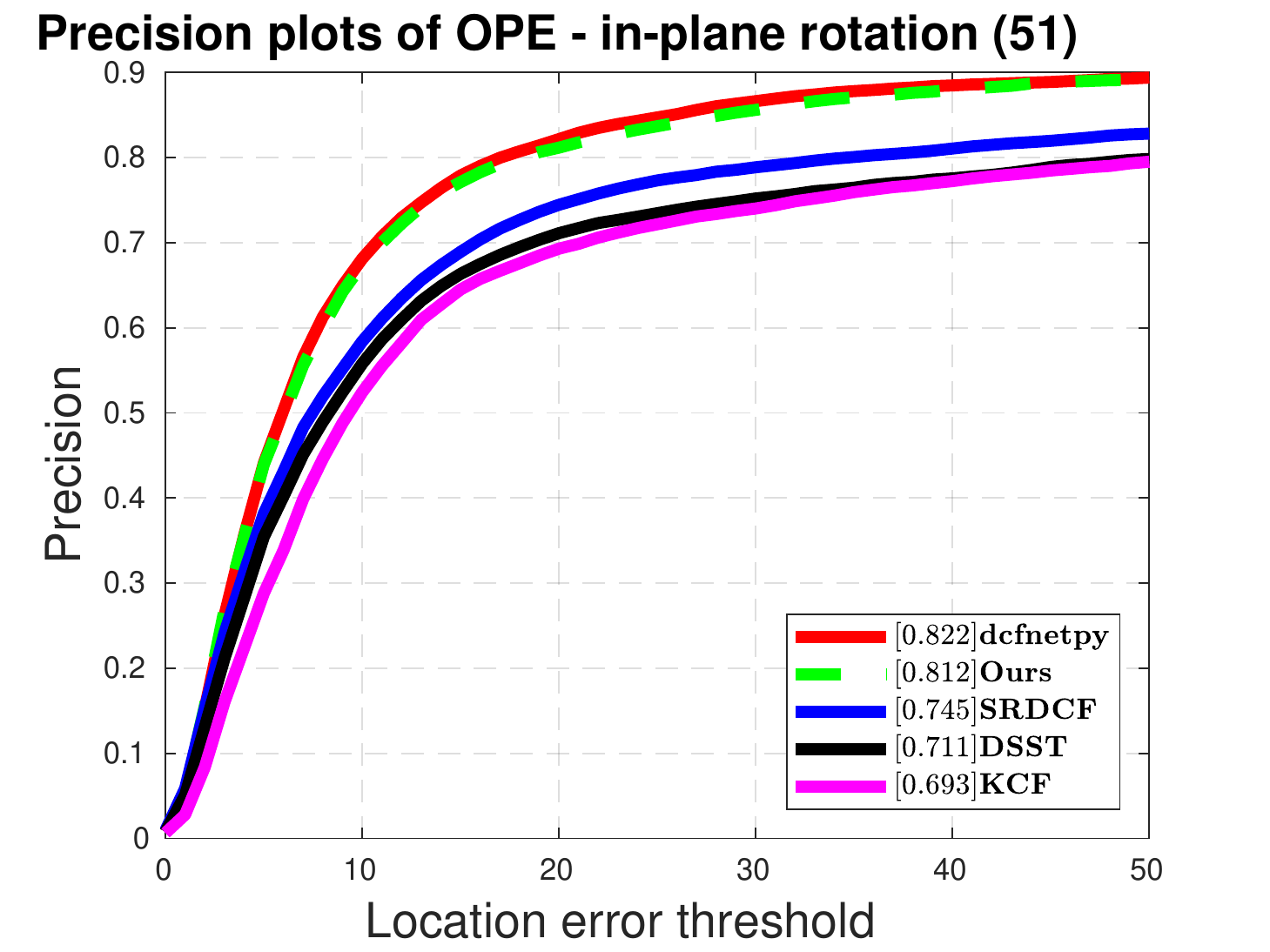}
\includegraphics[width=0.23\textwidth]{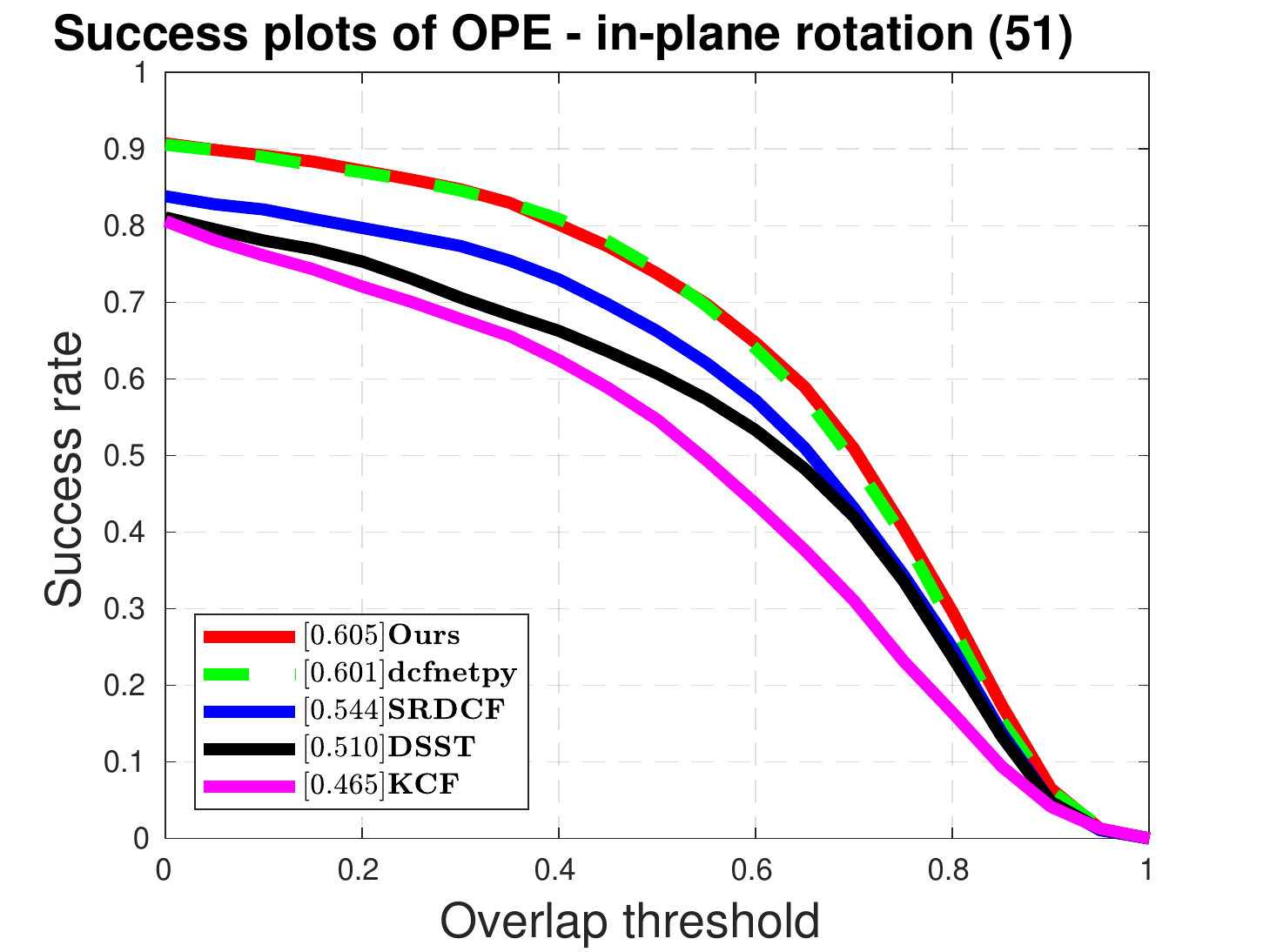}\\
\includegraphics[width=0.23\textwidth]{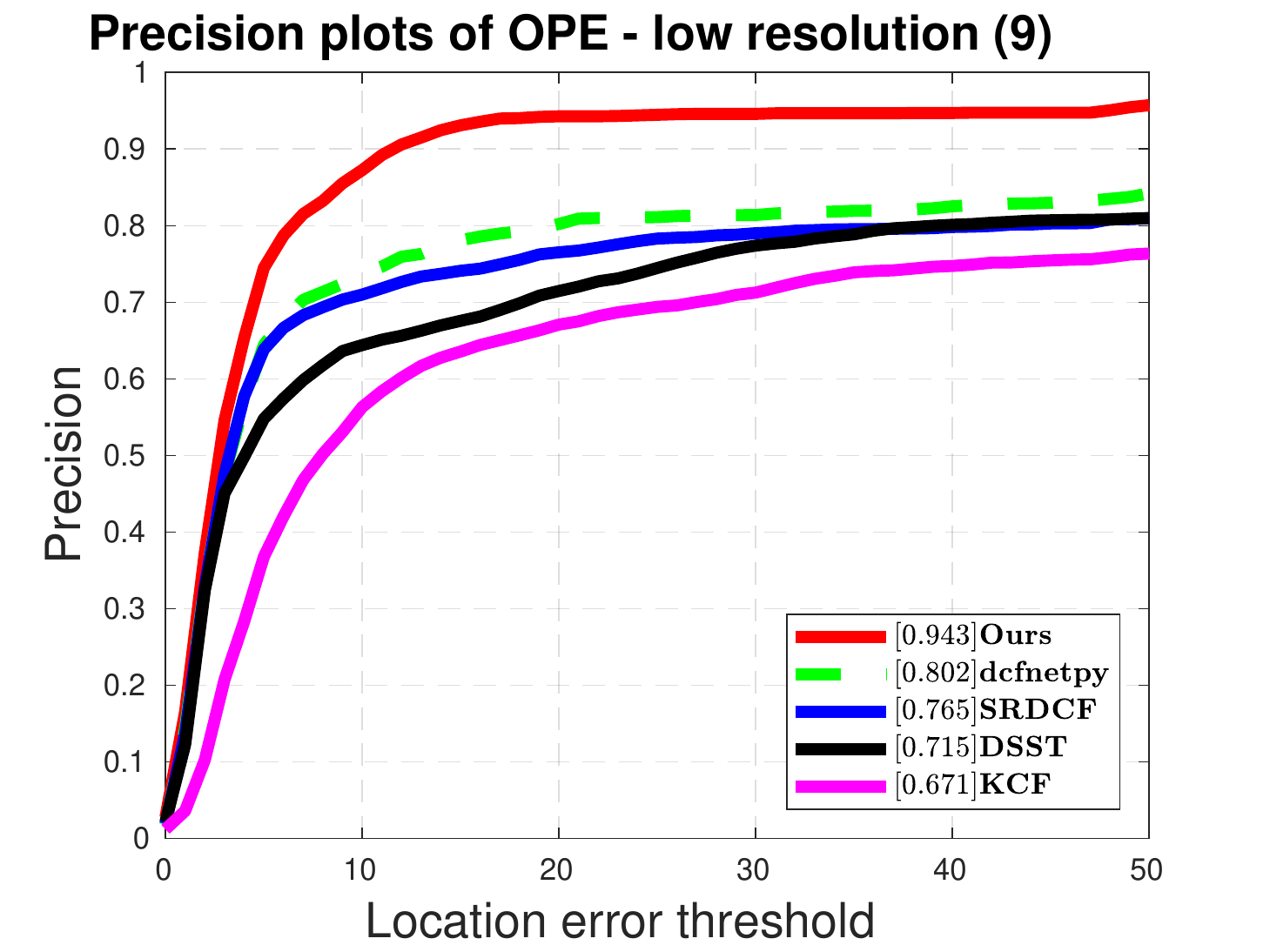}
\includegraphics[width=0.23\textwidth]{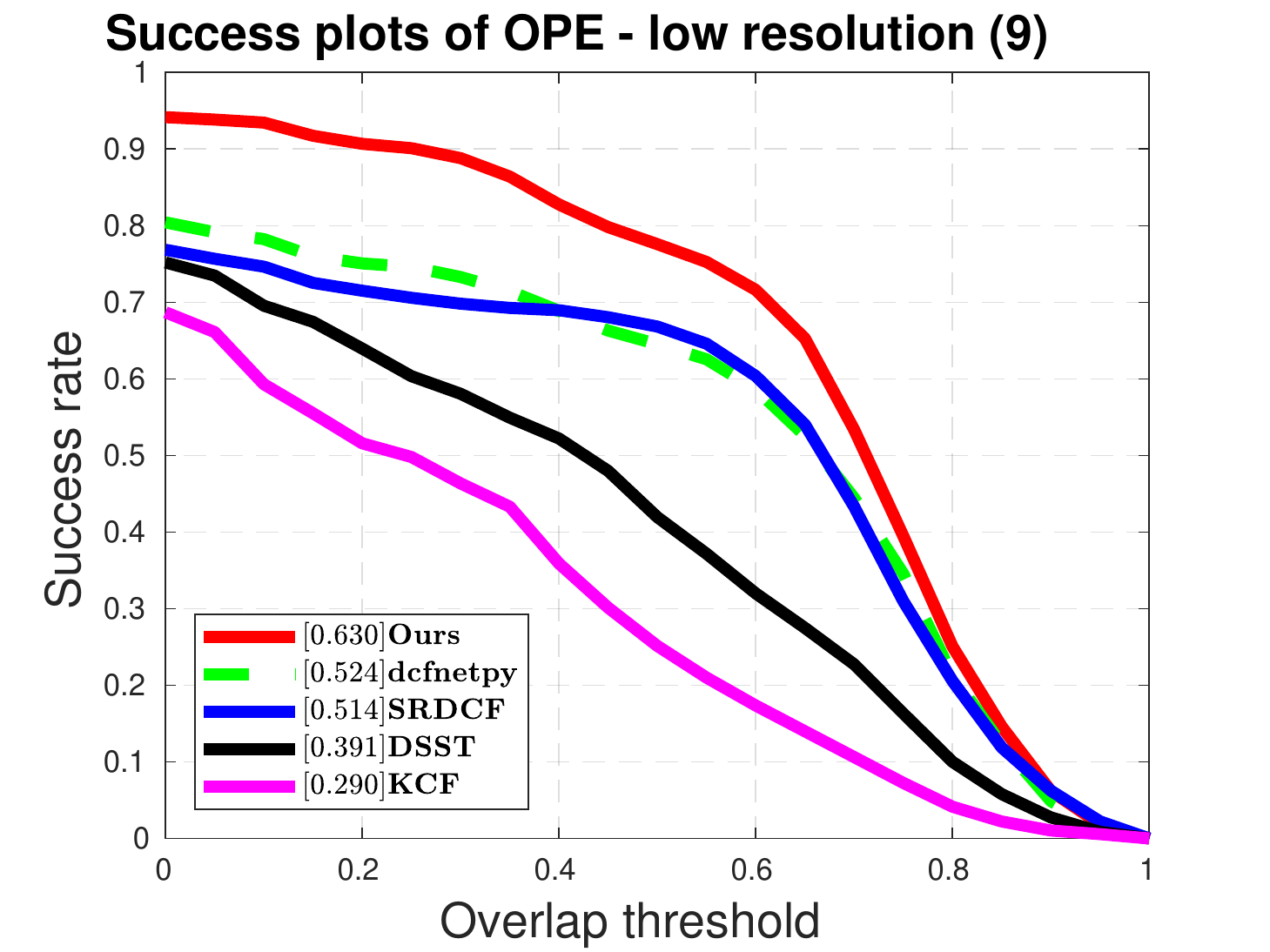}
\includegraphics[width=0.23\textwidth]{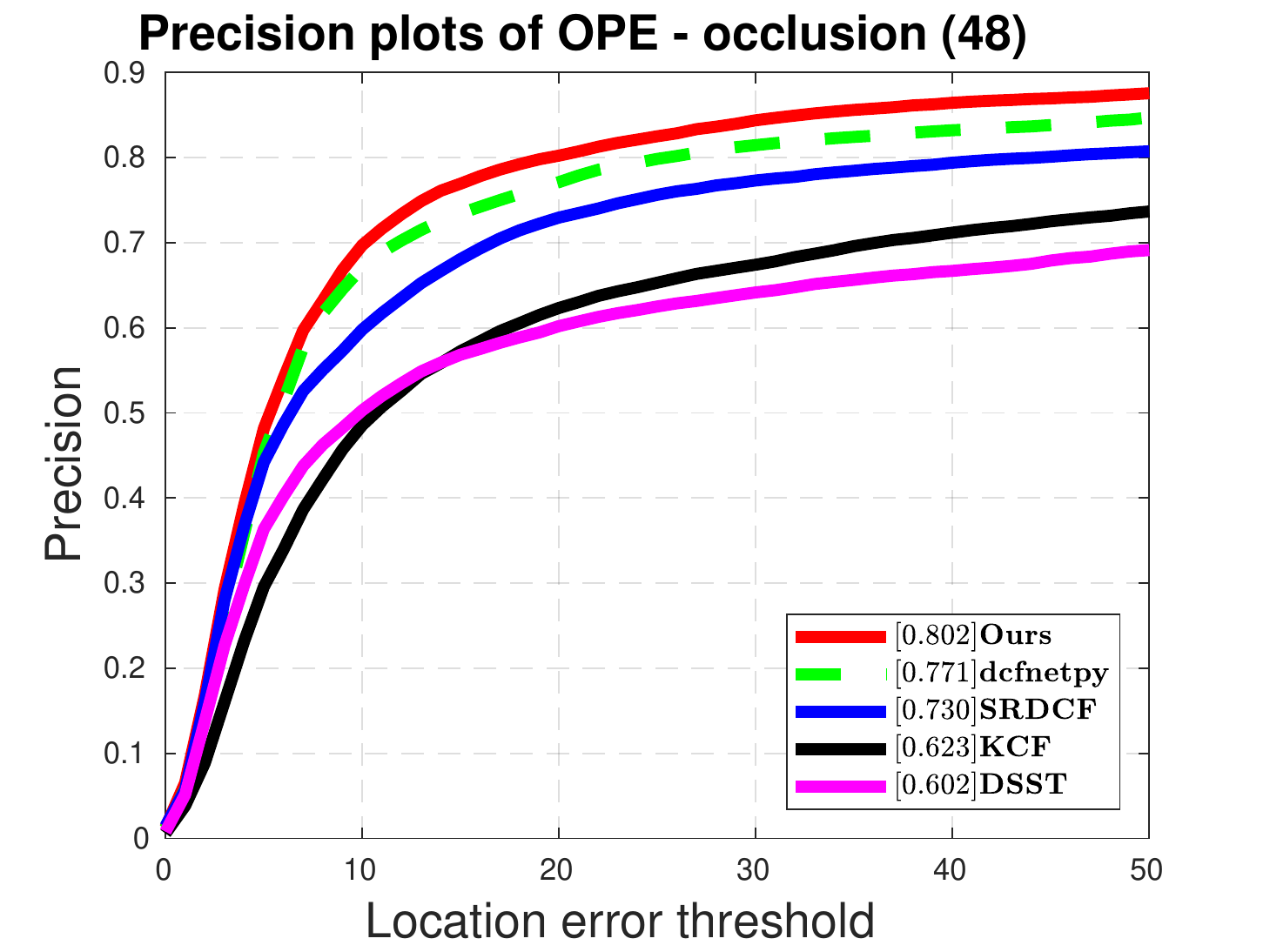}
\includegraphics[width=0.23\textwidth]{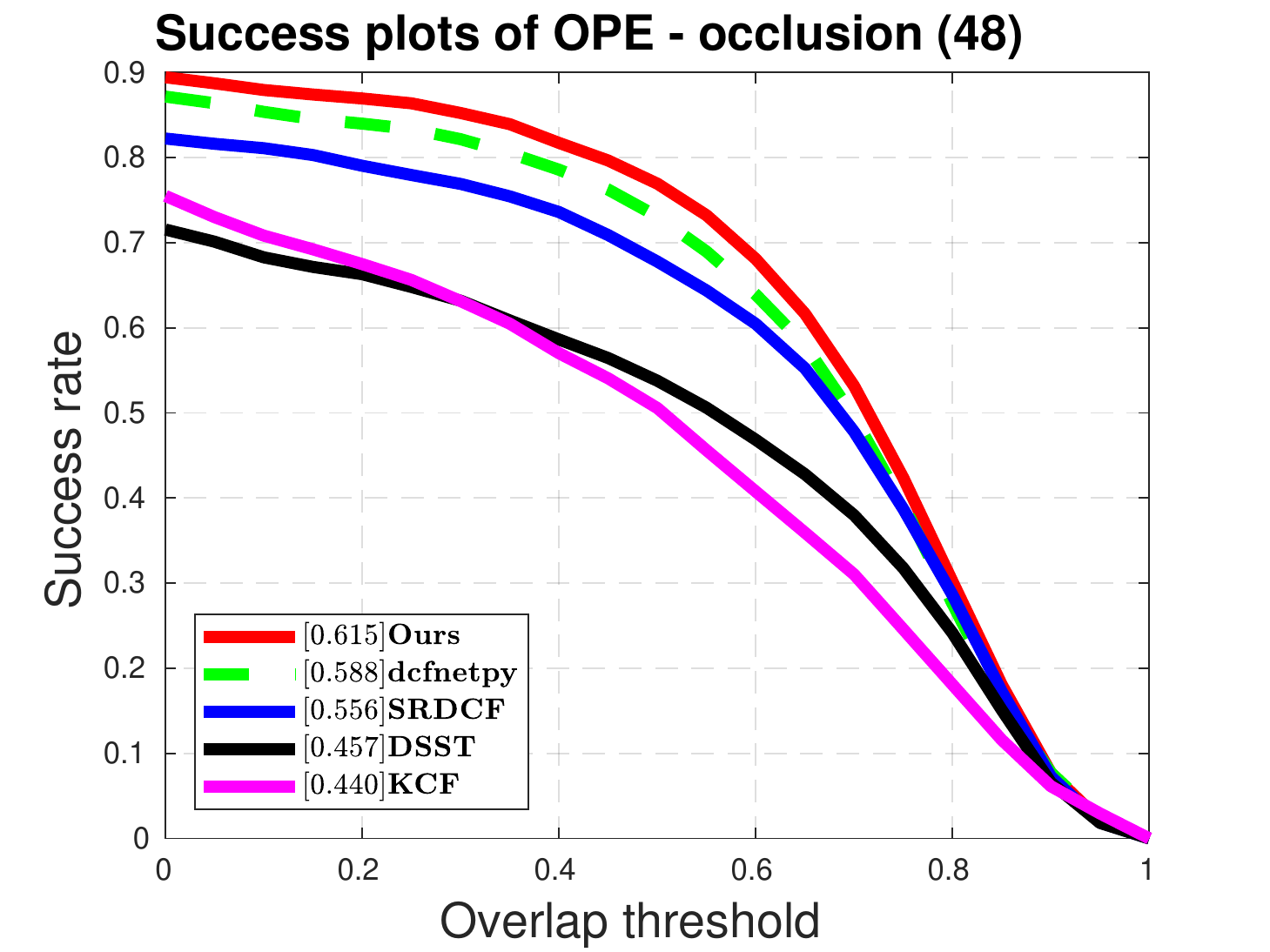}\\

\includegraphics[width=0.23\textwidth]{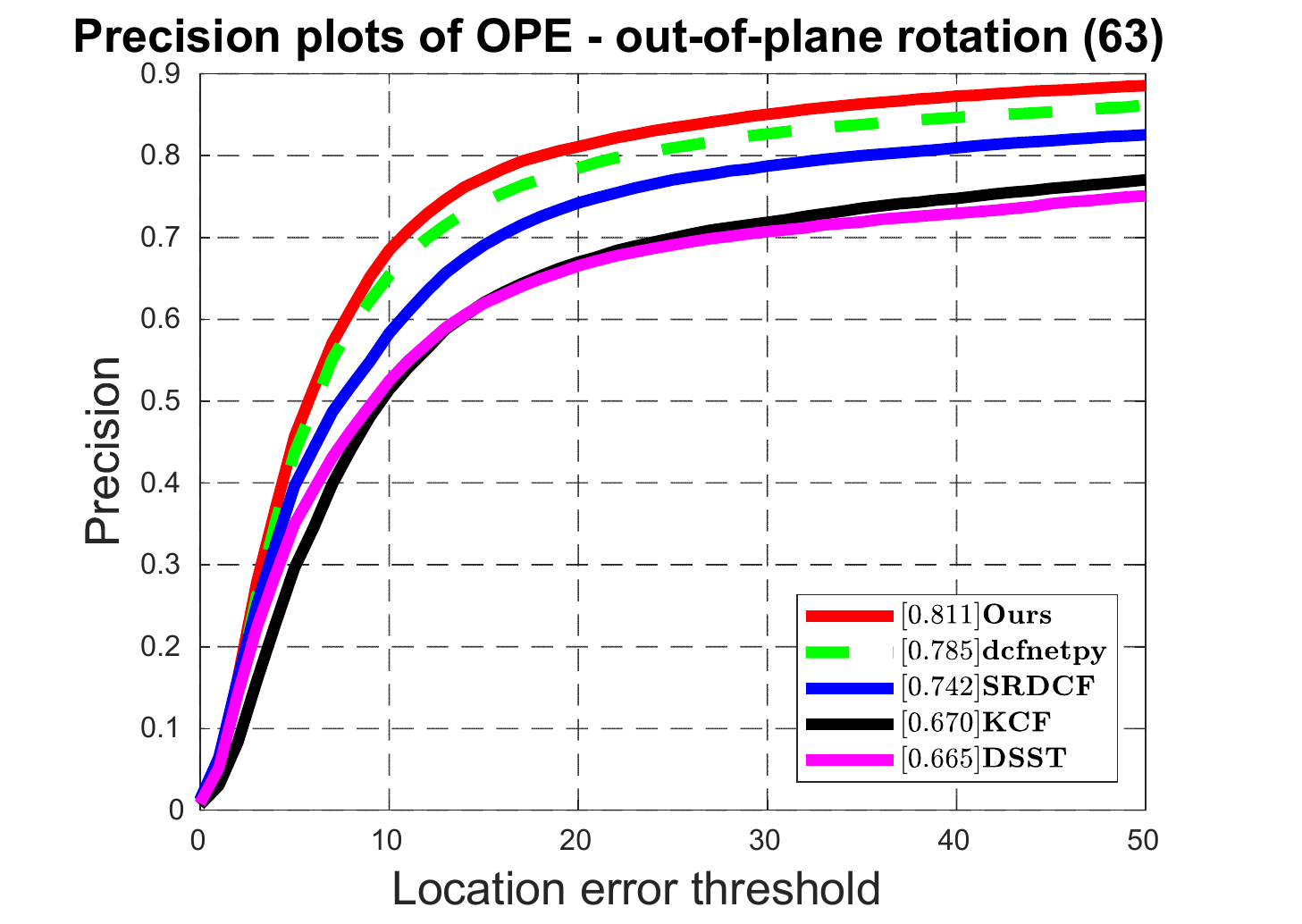}
\includegraphics[width=0.23\textwidth]{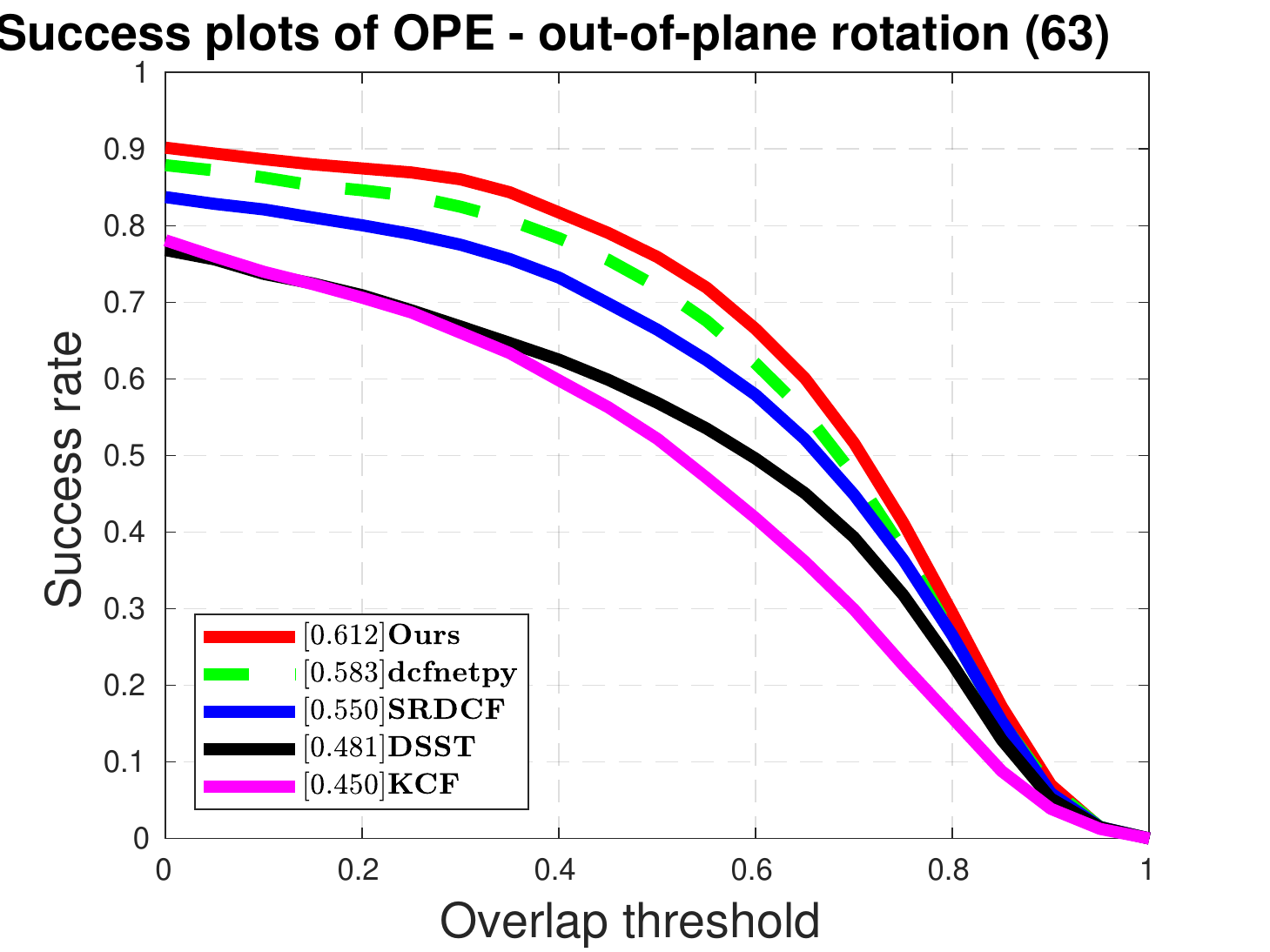}
\includegraphics[width=0.23\textwidth]{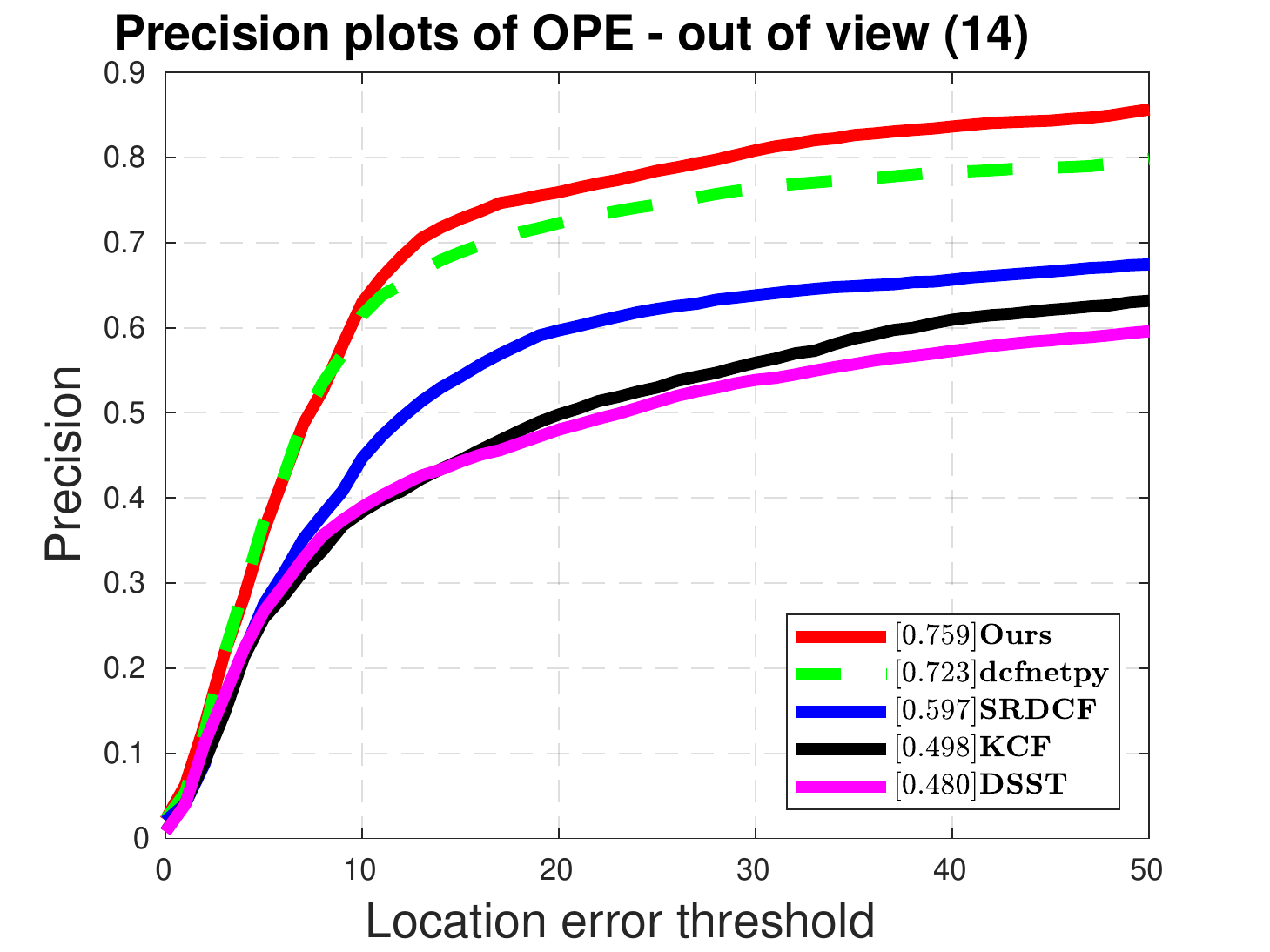}
\includegraphics[width=0.23\textwidth]{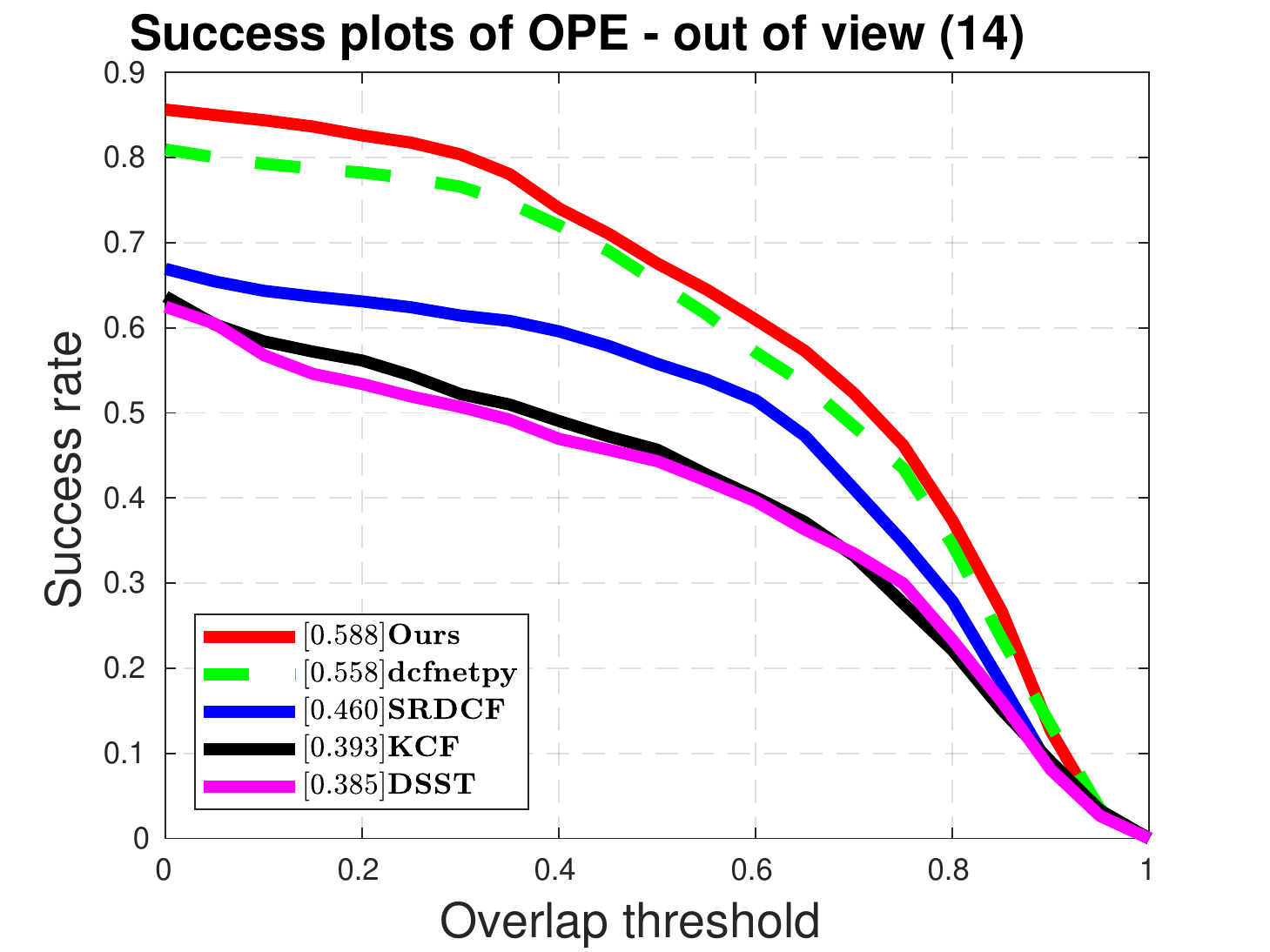} \\
\includegraphics[width=0.23\textwidth]{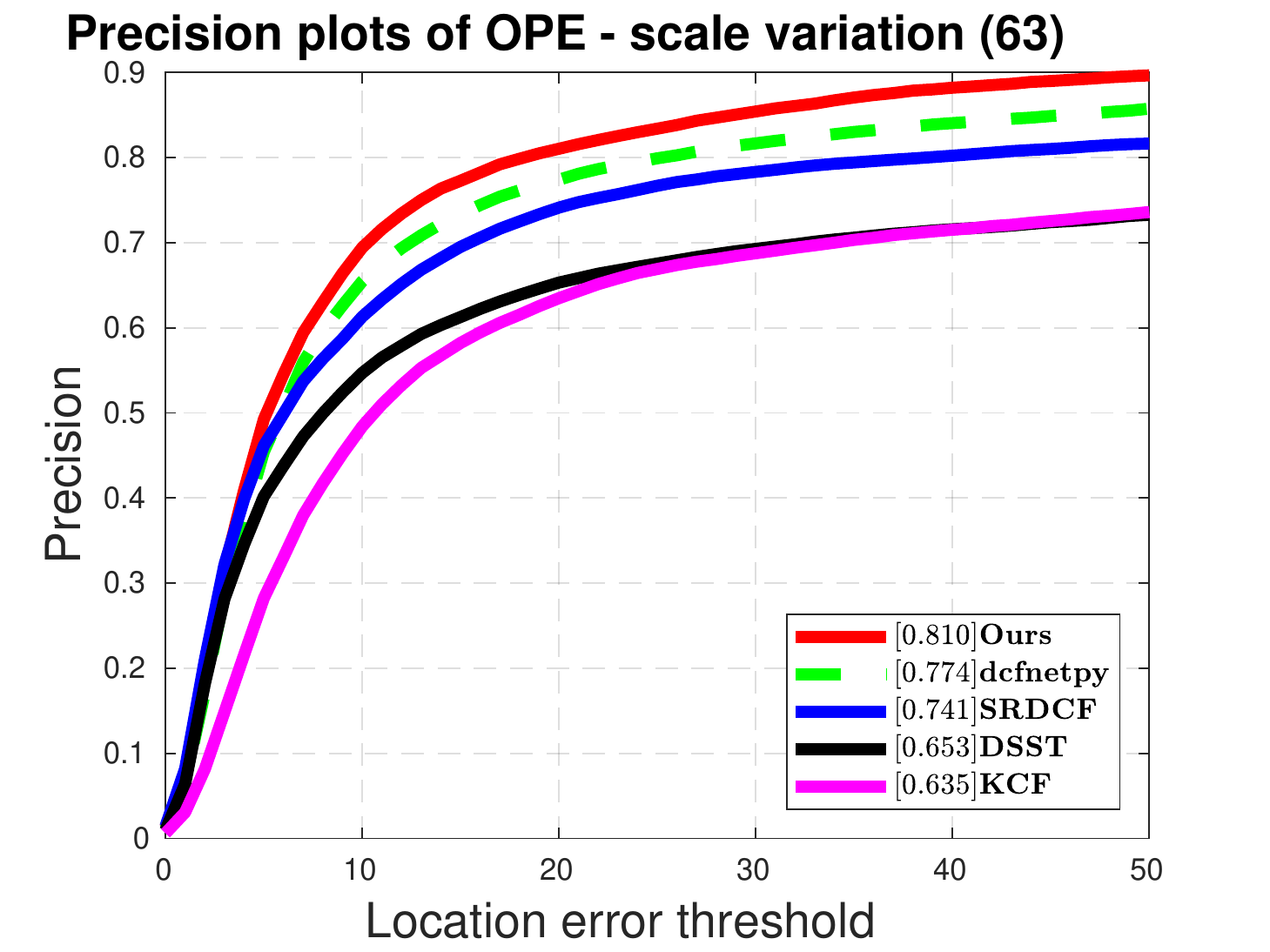}
\includegraphics[width=0.23\textwidth]{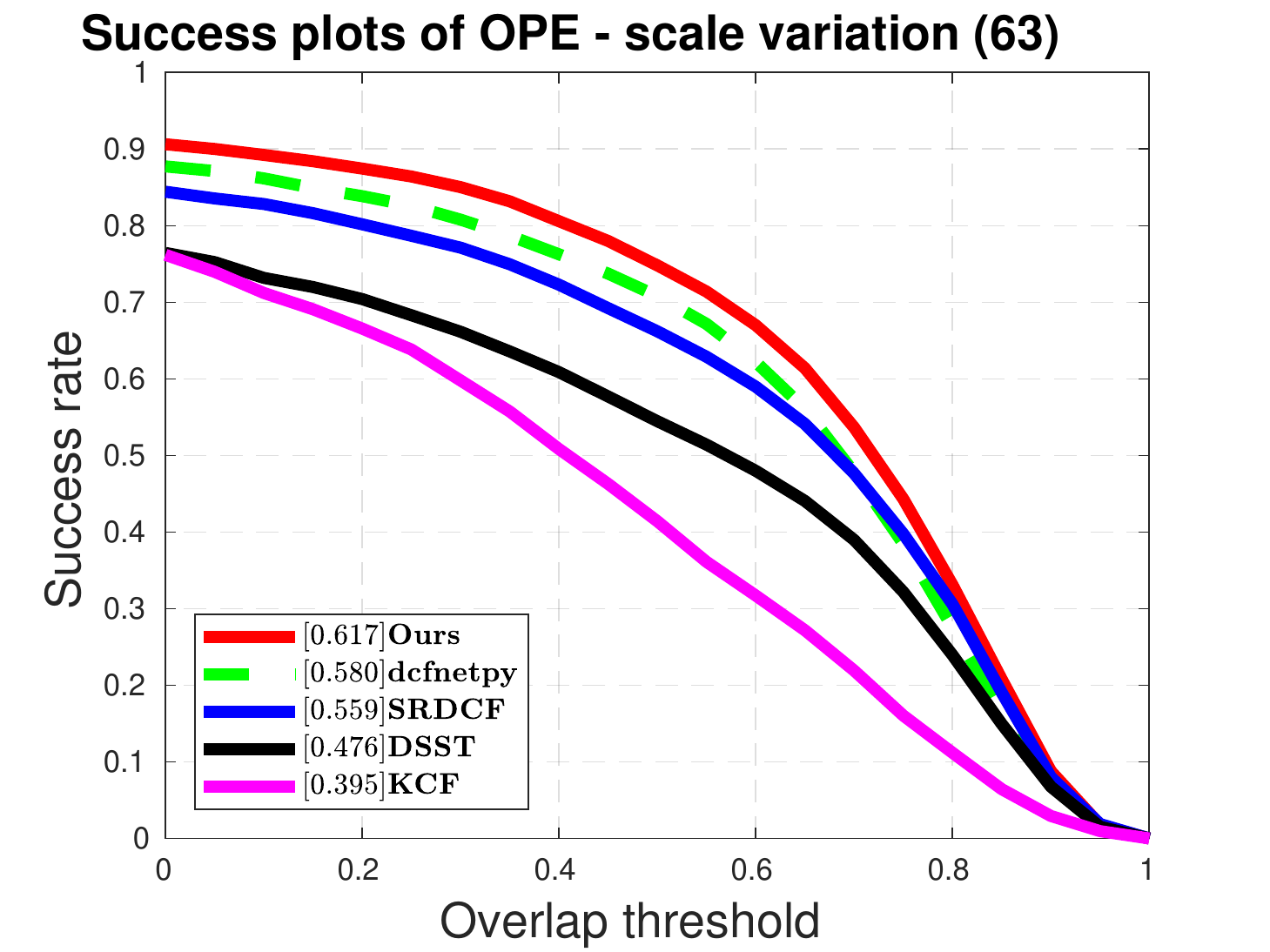}

\caption{The performance of 5 CF-based trackers for 11 attributes on OTB100, which contains 100 video sequences. \textit{dcfnetpy} is our python implementation of DCFNET. Our proposed model achieves higher success rate and precision compared with others.}
\label{attr}
\end{figure*}

\bibliography{mybibtex}
\bibliographystyle{IEEEtran}

\end{document}